# Holding AI to Account: Challenges for the Delivery of Trustworthy AI in Healthcare

Rob Procter[†]

Department of Computer Science, University of Warwick & Alan Turing Institute for Data Science and AI, rob.procter@warwick.ac.uk

Peter Tolmie

Information Systems and New Media, University of Siegen, peter.tolmie@uni-siegen.de

Mark Rouncefield

Information Systems and New Media, University of Siegen, mark.rouncefield@uni-siegen.de

The need for AI systems to provide explanations for their behaviour is now widely recognised as key to their adoption. In this paper, we examine the problem of trustworthy AI and explore what delivering this means in practice, with a focus on healthcare applications. Work in this area typically treats trustworthy AI as a problem of Human-Computer Interaction involving the individual user and an AI system. However, we argue here that this overlooks the important part played by organisational accountability in how people reason about and trust AI in socio-technical settings. To illustrate the importance of organisational accountability, we present findings from ethnographic studies of breast cancer screening and cancer treatment planning in multidisciplinary team meetings to show how participants made themselves accountable both to each other and to the organisations of which they are members. We use these findings to enrich existing understandings of the requirements for trustworthy AI and to outline some candidate solutions to the problems of making AI accountable both to individual users and organisationally. We conclude by outlining the implications of this for future work on the development of trustworthy AI, including ways in which our proposed solutions may be re-used in different application settings.

[†]Corresponding author

CCS CONCEPTS • Human-centered computing • Human computer interaction (HCI) • Computer-Supported Cooperative Work • Empirical studies in collaborative and social computing

**ACM Reference Format:**

## 1 INTRODUCTION

Trustworthy AI is an essential requirement for the effective, safe and ethical application of AI systems in decision-making support roles (HLEG 2019, Leslie 2019). AI systems are becoming an increasingly commonplace feature of organisational settings and are notably being promoted to support certain kinds of decision-making in healthcare organisations. In this paper, we argue that this means AI systems must be capable of providing accounts of their behaviour in ways that not only meet the requirements of individual users but that are organisationally appropriate (Cummings 2006). To do this, the paper draws upon a rich body of empirically-gathered materials to explore in detail how organisational settings may shape the requirements for trustworthy AI 'in the wild'. We put particular emphasis upon collaborative and organisational sense-making of the behaviour of AI systems and how members of organisations may interact with them.

A major challenge for trustworthy AI is that techniques have advanced in ways that have led the behaviour of AI systems to becoming increasingly 'black boxed'. This makes AI systems difficult – if not impossible – to understand by those that use them, and often even by those who build them (Pedreschi et al. 2018). As a consequence, much effort is now being devoted to developing ways of 'opening up' the black box (Carvalho et al. 2019, Du et al. 2019, Meteier et al. 2019, Mittelstadt et al. 2019) through the development of techniques for explainable AI (xAI) (Adadi et al. 2018, Gilpin et al. 2018). These aim to provide accounts of AI system behaviour that are transparent to – and interpretable by – their users. xAI techniques may be applied globally (i.e., to explain the behaviour of the system as a whole) or locally (i.e., to explain its behaviour for a specific input) (Guidotti et al. 2018). Current examples of xAI techniques include saliency maps, which visualise the most significant regions of input data for a prediction, and semantic disentanglement, which extracts from the underlying model high level features that are comprehensible by people (Henne et al. 2020). However, with some notable exceptions (e.g., Cai et al. 2019b, Ehsan et al. 2021), few studies have sought to assess the potential fit of these techniques with real-world requirements (Antoniadi et al. 2021). Doing this involves, first of all, a better understanding the socio-technical constitution of the organisational environments within which such systems might be deployed and how trust in the accountable character of those environments is currently achieved. Exploring the socio-technical challenges confronting the development of trustworthy AI in organisational settings is something that has so far been largely absent from the related literature (e.g., Yang et al. 2019, Liao et al. 2020, Ehsan et al. 2021, Glaser et al. 2021).

Healthcare is notable for high rates of failure in the adoption of digital innovations (Greenhalgh et al. 2017). Computer decision-support systems (CDSS) in clinical work are no exception, with many never progressing beyond the preclinical or pilot stage (Beede et al. 2020, Egermark et al. 2022, Oakden-Rayner et al. 2022). The causes are often complex, but some common factors can be identified. The first generation of CDSS were rule-based, which often proved too costly to develop in practice (Musen et al. 2014). The move to techniques based on belief networks and case-based reasoning helped to eliminate that obstacle, only to expose new problems. For example, Heathfield and Wyatt (1993) found that CDSSs were failing to address the real-world needs of clinicians, a problem that can be addressed by involving clinicians closely in the system design and development process. However, problems may still be encountered if it is assumed that a CDSS that has been designed to meet the needs of a specific group of clinicians will work as well in new clinical settings (Black et al. 2011, Nix et al. 2022, Oakden-Rayner et al. 2022). The emergence of deep learning-based techniques synonymous with AI has significantly extended the power and range of CDSS applications in healthcare but also brings new challenges. According to Yang et al. (2022), deep learning's lack of explainability is one of the main factors inhibiting adoption of this new generation of CDSS. Understanding just what it might take to make AI more explainable requires looking at how decision-making is done, not only at the level of the individual clinician but as a socially-situated practice (Aversa et al. 2018).

In this paper, we look at how socially-situated decision-making in healthcare is bound up with the organisational circumstances within which those decisions are taken, i.e., they are oriented to as organisationally accountable decisions. To do this, we examine in detail the practices that people follow to make themselves accountable, so that their co-workers can see their actions to be organisationally appropriate. Multiple studies have revealed how the timely and dependable completion of organisational work requires participants be able to make sense of each other's activities (Hartswood et al. 2007, Heath and Luff 1991, Hughes et al. 1994, Procter et al. 2006, Suchman et al. 2002). Participants 'do' being accountable as part of their everyday work, demonstrating competence in their role(s) and furnishing evidence of their trustworthiness. Moreover, providing evidence of trustworthiness is not a one-time act but must be continually reproduced in and through the course of people's daily activities (Clarke et al. 2006, Procter et al. 2022). Other studies of the explainability of AI and of how such explanations, or accounts, might serve as a foundation of trust in decision-making have largely left untouched the matter of how accounts are reflexively bound up with the organisational context within which they unfold. Our primary contribution here is therefore to bring into view the importance of understanding organisational accountability in relation to the delivery of explainable and trustworthy AI in healthcare.

In section 2, we begin our exploration of requirements for trustworthy AI by reviewing the literature on accounts and accountability. We examine in depth the nature of accounts and accountability practices, with a focus on organisational accountability and trust in professional work. It is important to note that accountability practices are shaped by the technologies in use and the ways in which these lead to specific forms of socio-material practice (e.g., Heath and Luff 1991). Introducing a new technology, such as AI, into an organisational setting may call for the reconfiguration of these practices but also of the technology itself so that organisational members can use it and/or adapt it in ways that best affords their accountability needs (Sellen & Harper 2003). Being organisationally accountable means that not only must AI systems furnish accounts that meet the needs of organisational members to make sense of their behaviour, but these accounts must also be compatible with how organisational members themselves manage being accountable to one another and to the organisation at large.

The safety critical demands of decision-making in healthcare make it a particularly perspicuous setting for exploring what organisational accountability involves. Section 3 presents three case studies drawn from a range of studies we have undertaken regarding breast cancer screening and treatment planning in multi-disciplinary team meetings. Through these, we explore what being accountable might amount to in relation to diagnostic work in healthcare. The findings are then used in section 4 to examine the challenges they present for data visualisation and HCI design and to propose some candidate solutions, before articulating some broader considerations related to auditing and regulatory frameworks for AI in healthcare. Finally, based on these we formulate the implications of this material for an ongoing programme of work in section 5.

## 2 AN OVERVIEW OF ACCOUNTS AND ACCOUNTABILITY

In this section we summarise key points from the literature regarding accounts and people's everyday accountability practices. We explore how accounts and accountability have distinct characteristics that are of enormous significance for how they are handled by people. One part of this relates to the nature of accounts themselves and whether they might be considered *formal*, *situated* or *natural*. The latter two of these are tightly interrelated but can have different outcomes. Another consideration is how accounts and accountability relate to the context within which they are embedded. Of particular relevance for this paper is how accounts have an indexical and reflexive relationship with organisational settings, i.e., organisational accountability. This relates to the way in which people orient to understood organisational structures, requirements and imperatives when managing their accountability, not just to each other, but to the organisation as well. Here, other members of the same cohort are understood to be organisational incumbents who reason about the organisation in similar ways.

Another consideration is how accountability can be seen to have *grammatical* characteristics (Coulter 1983, 1989). What we mean by this is that what people do and how they do it has an accountable order and way of being put together. This can be a spatial and temporal matter, e.g.: certain things get done in certain places and not in others; things may have a certain sequence; things may get done at certain times and not at others; etc. It can also be about coherence or co-occurrence, e.g.: things may be describable in one way, but not in another; some things may be seen to go together, other things not.

We conclude this section by looking at how notions of 'trust' and accountability can be seen to be related and what this might mean in the context of trustworthy AI.

**2.1 A typology of accounts**

As noted above, accounts can be considered as formal, situated or natural. This typology has significance when examining the implications of introducing new technologies and systems into the organisational workplace.

Formal accounts are accounts of action that are situation independent, such as stories, scripts, minutes, written explanations, scripted commentaries, etc. Although they may be engaged with in different ways in different situations, they are typically understood to be reproducible across a range of different circumstances and thus, in some sense, generic (there is more to the matter than this, but we shall return to it below). The disadvantage of formal accounts is that their prescriptive character can make it difficult for them to be adapted to fit the specific requirements of the situation in which they are produced.

Situated accounts, by contrast, are accounts that are tailored to circumstance. Fundamentally, they are possessed of three interactionally distinct modes of production. First, they may be produced as a response to being 'called to account'. If the reasons why some course of action has been pursued by one party are not evident to another party, the latter may request an explanation, i.e., an 'account'. This request typically takes the form of direct questions, e.g., 'What are you doing?', etc. Second, a party who has pursued some course of action may suspect it was not understood correctly by others around them. They may then produce a pre-emptive account to explain their actions, so that they are never actually called to account. A third form of situated account is where the explanation of some course of action is produced in company with the course of action itself. It may be pre-scripted to some degree, but can be adapted to the moment of production, enabling it to accommodate contingency and the particular recipients to hand.

Two key characteristics of all situated accounts are: i) they are recipient designed (Sacks 1992), i.e., the exact way they are formulated is tailored to their recipient, so two different individuals may receive different accounts, depending upon matters such as their competence and their relationship with the party producing the account; ii) they are tailored to the circumstances of their production, i.e., they are positioned appropriately within an ongoing sequence of interaction and demonstrate due recognition of the ecological and semantic circumstances.

Centrally, situated accounts are able to draw upon the circumstances of production as a resource (i.e., they display a reflexive and indexical relationship with the situation). One advantage of situated accounts is that by being tailored to fit the immediate need they are much more likely to be readily understood. Another advantage is that they are open to repair. In other words, a recipient's failure to understand can be recognised by the account-giver and they can then reformulate the account until it is understood. Some disadvantages of situated accounts (at least from the point of view of AI systems design) are: the need for an account has to be recognised in the first place; the account-giver has to be able to grasp the local context; and the account-giver has to be able to make the account fit the need, i.e., they have to not only to be capable of grasping the local situation but also have the competence to be able understand how to respond appropriately.

As people conduct their everyday affairs, they take the ordinary, readily explicable character of the world around them for granted. This taking for granted of the explicable character of phenomena and their constituent features can be termed as treating them as 'naturally accountable' (Garfinkel 1967). Implicit to this is a notion of 'trust', which is "a background condition for mutually intelligible action" (Watson 2009: 476). Trust is a key component of people's 'mutual commitment' to the 'rules of engagement', where "all parties to the interaction must understand that they are engaged in the same practice, must be competent to perform the practice, must actually perform competently and assume this also of the others." (Watson 2009: 475). This forms the intersubjective grounds of everyday action and its intelligibility, i.e., "We trust in other parties' ability and motivation to make similar sense of a situation, using similar sense-making methods and instruments." (Watson 2009: 481). Of course, occasions do arise where our background expectations are challenged and this is when the production of situated accounts occurs, as described above.

Clearly, the advantage of natural accountability is that there is no need for an overt account and all courses of action are, for the larger part, mutually intelligible to those engaged in them or witnessing them. This also means that trust is implicit because everyone understands one another to be oriented to a shared set of background expectations. Natural accountability calls for having the competence to see the world in the same way as those with whom you are interacting, and they have to see you as being possessed of that competence. This may be a 'vulgar competence' (Garfinkel 1967), i.e., something that is shared by most other people, or it can be a competence shared by members of a more specific cohort, for example, clinicians.

There are relatively few direct discussions of how people handle accountability and trust in the digital domain. One persistent and growing thread of interest relates to the accountability of systems and how effectively systems might make visible to their users 'what they are doing', i.e., give some account of their actions. This notion can be traced back to the work of Dourish (1993, 1997, 2001a), Button and Dourish (1996), and Belloti and Edwards (2001). It continues to be of importance for the design of trustworthy IT systems (Eriksen 2002) but, as we have noted above, much of the recent discussion about it has been framed within a growing body of work on the explainability (or otherwise) of AI algorithms (e.g., Abdul et al. 2018).

While trustworthiness may be seen to be a generic requirement of IT systems, the challenges of meeting it may vary significantly with the kind of system in question. In some cases, users may treat trustworthiness as an accountable matter in *very specific circumstances only* (Dourish 1997, Anderson et al. 2003), such as when an Internet connection unexpectedly slows or stops working. In other cases, users may treat trustworthiness as an accountable matter *routinely*, such as when a recommender system suggests products to an online shopper, or an AI system presents a candidate diagnosis to a clinician. However, what distinguishes the provision of accounts by recommender systems, for example, from those by an AI system in healthcare is that in the latter, the AI account will be a material resource for an account *to be provided by healthcare professionals* for whatever action they make. This is of fundamental importance as it is the trustworthiness and organisational accountability of the healthcare professional that is at stake (Nix et al. 2022).

With this in mind, one question we will be seeking to address is what current approaches to xAI are actually addressing: formal accountability, situated accountability, or natural accountability. The first of these is clearly the easiest to accomplish; the last might be considered the 'ideal'.

In summary, everyday life is socially organized and this social order constitutes a moral order. This order is the basis upon which the very need for an explicit account or otherwise is founded. It provides for seeing what needs to be done or left alone, for seeing what one's obligations might be and for seeing what might or might not be reasonably expected of others. The latter is of particular concern in organisational contexts and it is a foundational characteristic of healthcare settings. As these form the backdrop of the case studies in this paper, we will look next at the topic of organisational accountability, which is fundamental to establishing and sustaining trust in professional work. If AI systems do not provide the resources to support this, then this trust will be undermined.

**2.2 Organisational accountability**

While notions of organisational accountability are tightly bound up with the preceding considerations, they differ in the way they focus upon the relationship between people and organisational structures, requirements and imperatives. Here, the intersubjective aspects outlined above are premised upon assumptions that other members of the same organisational cohort consider themselves to be accountable to the same organisational concerns in much the same way.

All aspects of organisational life, from management protocols to technologies are subject to the same concerns. This underpins how organisations are constituted as social phenomena and, at heart, it makes trustworthy AI first and foremost a socio-technical problem. There are a number of treatments of this topic in the literature.

Perhaps the earliest ethnomethodological discussion of organisational accountability can be found in Garfinkel's "Good organizational reasons for 'bad' clinical records" (Garfinkel 1967: 186-207). This relates to a study that Garfinkel and Bittner undertook of the nature of clinical records in a psychiatric clinic in California in the 1950s. Garfinkel and Bittner's remarks are primarily addressed to the absence of seemingly important bits of information from the large majority of the files they inspected. As they tried to unravel and reconcile these absences, they came to see the problem as one of dealing with 'normal, natural troubles', which "occur because clinic persons, as self-reporters, actively seek to act in compliance with rules of the clinic's operating procedures that for them and from their point of view are more or less taken for granted as right ways of doing things" (Garfinkel 1967: 191). In his explication of the issues, Garfinkel draws a sharp distinction between the accountability of clinic records as *actuarial records*, in relation to which they were clearly wanting, and as records "of *a therapeutic contract* between the clinic as a medico-legal enterprise and the patient" (Garfinkel 1967: 198 (original italics)), in relation to which they were perfectly adequate when certain assumptions were being made. Thus, Garfinkel suggests that:

> "In order to read the folder's contents without incongruity a clinic member must expect of himself, expect of other clinic members, and expect that as he expects of other clinic members they expect him to know and to use a knowledge (1) of particular persons to whom the record refers, (2) of persons who contributed to the record, (3) of the clinic's actual organization and operating procedures at the time the folder's document are being consulted, (4) of a mutual history with other persons – patients and clinic members – and (5) of clinic procedures, including procedures for reading a record, as these procedures involved the patient and the clinic members. In the service of present interests, he uses such knowledge to assemble from the folder's items a documented representation of the relationship." (Garfinkel 1967: 206).

It can be seen within this an assumption of the natural accountability of clinic records to other members of the same organisation who can grasp the same intersubjective point of view. This goes to the heart of organisational accountability: it is premised upon a presumption of the adequacy of one's actions, including the production of records, not to just anyone at all, but rather to just anyone who has available to them an understanding of an organisation's operating procedures and 'the right way of doing things' as a member of that organisation. This is evidently a substantial elaboration of what was said in the previous section regarding formal accounts. The significant thing to grasp is the distinction Garfinkel makes between what might be thought of as an actuarial, generic use of formal accounts and the ways in which they are actually generated and used in practice. We return to some of these points in the discussion because they constitute a vital backbone to much of the reasoning visible in the case studies we present below.

There have been various developments of the discussion of accounts in Garfinkel and Bittner's original study of clinical records. Bittner, for instance, moved beyond this to write a seminal paper on organisational accountability that took to task organisational theory at the time (Bittner 1965). Bittner's key concern was to illustrate how sociological discussions of how organisations are constituted managed to miss how organisations are oriented to by organisational members in everyday practice. His argument revolves around two principal concerns, to which we shall return later. One of these is what he termed 'gambits of compliance'. This is how organisational members find ways to account for their actions in terms of what just any member knows about that organisation and its interests. Gambits of compliance also serve to make manifest organisational members' understanding of the organisation they inhabit. Thus, they make available to the analyst an intersubjectively-established, common-sense understanding of what an organisation amounts to.

Bittner's second, strongly related point, is that organisational members orient to what they understand to be the 'stylistic unity' of an organisation. This refers to how organisations, despite their potential scale and diversity, are still taken by members to be a sphere of 'concerted action'. That is, despite the presence of specific rules for specific matters to which members might consider themselves accountable, they also orient to a sense of there being an overall interest or 'reproducible theme' to which an organisation adheres. Bittner further notes how organisational members will use the stylistic unity of an organisation as a source of 'corroborative reference', whereby their overall understanding of an organisation's purpose can be invoked as an account for what might otherwise seem fragmentary or contradictory. In a more recent development of these themes, Tolmie and Rouncefield (2016) have examined the visible exercise of organisational acumen by members of an organisation in relation to how they prioritise activities, demonstrate adherence to organisational policy, and preserve a sense of organisational consistency.

When it comes to how notions of organisational accountability have been used in the digital domain, it is important to make a distinction between how members might be seen to orient to formally constituted records and procedures, and how members can be seen to deliver organisationally appropriate accounts, for instance, through gambits of compliance. A classic example of the former is Suchman's discussion of how plans stand not as instructions but rather resources for situated action (Suchman 1987). Examples of the latter include Dourish's discussion of the use of workflow technologies (Dourish 2001b) and Martin et al.'s study of the development of electronic patient records (Martin et al. 2009).

**2.3 The grammatical constitution of accounts**

Wrapped into all of the previous considerations are the ways in which accounts and accountability may be seen to have grammatical characteristics, i.e., they have a recognisable order and are assembled in certain kinds of ways. Accounts and accountability are tied up with: 1) spatial concerns – things get done in certain places and not others and are accountable in those terms; 2) temporal concerns, where, a) courses of action are in a temporal order that is often sequenced in a certain way, and b) things are done at certain times and not others; 3) coherence concerns, where, a) things may be describable in one way, but not in another, b) some things may be seen to go together, others not; c) the occurrence of one thing may strongly implicate the co-occurrence of another, and d) relatedly, the description of one thing in a certain way may strongly implicate the description of other co-situated elements in a certain way.

Spatial accountability is a pervasive feature of our world. It takes little effort to see what kinds of activity are considered naturally accountable or otherwise in offices and factories, and so on. The sequential organisation of conversation and resulting patterns of accountability were first described in detail by Sacks et al. (1978). The notion of extending this sequential view of action through words to action has been articulated by Coulter in terms of 'grammars of action' (Coulter 1983, 1989). It will be seen in our description of multidisciplinary meetings in a breast cancer clinic that such meetings have a powerfully implicative sequential order that provides for a range of important outcomes. Another accountable aspect of temporality is the timing and duration of different things. Clearly, just when certain activities are completed and just how long they take can be of significant concern in organisational settings. This interest in temporal accountability has been previously described in several production environments (Button & Harper 1995, Button & Sharrock 1997), with it being potentially consequential for the management of divisions of labour.

The describability of certain things in certain ways, and the situated grounds of such descriptions, is of particular pertinence for this paper. There are ways in which appropriate descriptions are assembled on the basis of appropriate evidence that are tightly bound up with how things are seen in commonly oriented-to ways amongst specific cohorts. This has been a recurrent topic in ethnomethodological studies of expert practice, ranging from Garfinkel et al.'s early paper on the optical discovery of a pulsar (Garfinkel et al. 1981), through Goodwin's analysis of professional vision (1994), to more recent studies of work with mammograms (Hartswood et al. 2002a, Hartswood et al. 2007). With regard to the latter, Slack et al. (2010) unpacked the specific kinds of 'professional vision' involved in diagnostic work in terms of what might be called 'geographies' or 'topologies of suspicion', where it is an assembly of anomalous features that plays into what may be seen as a potentially carcinogenic mass or lesion. This reasoning about what things go together can also be seen to extend to recognition of what activities go together. Crabtree and Rodden (2004) note that certain activities can be seen to cluster in certain spaces in homes, with objects being placed within those locations actively promoting and supporting coordination between inhabitants (e.g., by positioning letters in certain places). This kind of concern is obviously of importance in workplace settings and more recent work has taken a similar view of how certain constellations of things may actively provide for reasoning and account (Anderson 2017).

There is one further 'grammatical' consideration of accountability of relevance to this paper that needs to be mentioned. Sacks (1992) pointed out that there are certain categories we are predisposed to hear as going together, such as mothers and babies, drivers and passengers, doctors and patients. Upon the basis of this, we regularly make ordinary and naturally accountable assumptions. Sacks termed such co-occurrent descriptors as membership categorization devices (MCDs). Understanding how appropriate assumptions of co-occurrence might occur is one of the challenges confronting trustworthy AI, when it comes to being able to assemble what are seen to be reasonable accounts in specific circumstances.

**2.4 Trust and accountability in the context of AI**

As a final point for consideration, in our discussion of natural accountability, we pointed to the interrelationship between accountability and trust (Watson 2009). Something we want to emphasise here is the particular importance of this in relation to the prospective use of AI systems to support decision-making in safety critical circumstances. As a broad topic, the relationship between trust and safety critical systems has generated a substantial literature (Albayram et al. 2019, Coskun and Grabowski 2004, Fenton et al. 1998, Johnson 2002, Mentler et al. 2016). Here, we want to bring to the fore a specific concern: if, as Garfinkel (1967) and Watson (2009) suggest, trust turns upon the intersubjective grounds of accountability, what will it take for an AI decision-support system to be trusted?

There is a similar substantial literature concerning improvements in AI and its role in decision-making, and associated ideas such as trust, accountability, and explainability (Knowles et al. 2014, 2015, Cai et al. 2019a, 2019b, Smith 2021, Kaur et al. 2022). Early work on the use of AI in detecting breast cancer (Hartswood et al. 2003) points to the importance of 'repair', of how clinicians are required to use their collaborative 'professional vision', i.e., "socially organised ways of seeing and understanding events that are answerable to the interests of a particular group" (Goodwin 1994) to make sense of the outputs of the technology:

> "That a mammogram feature or a prompt is there is not, of itself, constitutive of a lesion or other accountable thing, it must be worked up through these embodied practices and ratified in the professional domain of scrutiny. The machine knows nothing of what it is to be a competent, professional reader and what it is to look for features in a mammogram beyond its algorithms – that is self-evident – and the reader must 'repair' what the machine shows, making it accountable in and through their professional vision." (Hartswood et al. 2003)

More recent work has reiterated these findings, and their importance, identifying a failure to consider typical HCI issues, to explain system use or capability or its contribution to existing collaborative practices that might make it fit in to everyday work, or become 'unremarkable' (Yang et al. 2019). For example, McKinney et al. (2020) outline experimental evidence of statistical improvements in decisions and outcomes from using AI in breast screening but with little concern about how these might impact on the actual working practice and everyday experience of radiologists. In contrast, Smith (2021) highlights a range of interlocking and contradictory issues concerning accountability, responsibility and transparency in the use of AI in clinical decision making. Similarly, Henriksen et al. (2021) suggest the 'inherent' opacity of deep learning-based AI systems reduces the possibilities for accountability and trust, arguing for a move towards 'explainability' of AI systems. Abdul et al. (2018) stress the importance of understanding how the context of use impacts on requirements for explainability. Finally, Ehsan et al. (2021) argue for the need for critical reflection and a movement away from algorithm-centred approaches towards more social transparency in AI systems:

> "Implicit in AI systems are human-AI assemblages. Most consequential AI systems are deeply embedded in socio-organizational tapestries in which groups of humans interact with it, going beyond a 1-1 human-AI interaction paradigm. Given this understanding, we might ask: if both AI systems and explanations are socially situated, then why are we not requiring incorporation of the social aspects when we conceptualize explainability in AI systems?" (Ibid)

For us a telling aspect of this argument is the emphasis on 'socio-organisational tapestries', since it is exactly this organisational aspect of accountability that features so strongly in our own research into healthcare technologies and innovations.

The new generation of AI systems have a diverse range of potential applications in healthcare decision-making, including diagnosis (Bohr & Memarzadeh 2020). The roles they may be assigned to in diagnostic processes, for example, assisting or replacing human expertise (Keane & Topol 2018), remains an open question, one that will require thorough evaluation of performance, together with proof of meeting any regulatory standards (Arora 2020) and ethical principles that apply (Guan 2019). However, regardless of role, AI systems will need to be capable of being accountable for their decisions in ways that healthcare professionals can trust. Some studies, such as Wang et al. (2019), have approached these questions from a decision theory perspective. Other studies have followed an empirical approach based on interviews with clinicians (Yang et al. 2019, Liao et al. 2020) or scenario-based methods (Ehsan et al. 2021). What is lacking is an empirically-based exploration of what it means for AI to be accountable 'in the wild' and this is what we aim in this paper to begin to address by presenting evidence from our previous field studies of accounts and accountability practices in healthcare.

**3 CASE STUDIES OF ACCOUNTS AND ACCOUNTABILITY**

**3.1 Rationale**

As noted above, accounts make people's reasoning about particular courses of action manifest. People assume that the reasoning exhibited in their own and other people's accounts is appropriate for the organisational context within which they are situated. This is argued by Watson (2009) to be the cornerstone of intersubjectivity because accounts trade upon background expectations that just anyone within a particular cohort will bring to bear. So, examining accounts and explanations put forward by members of an organisation gives insight into the nature of accountability in that organisation and what it will take to support it or otherwise.

The safety critical demands of decision-making in healthcare, together with it being a very active domain for the application of AI, make healthcare an excellent choice of setting in which to explore the nature of organisational accountability and the design of trustworthy AI systems. We have undertaken many studies of technological innovation in healthcare settings over a span of some 20 years. In order to explicate in detail the kinds of accountable, expert reasoning present in healthcare settings, we have selected from this body of work a series of studies relating to the diagnosis and treatment of breast cancer. By presenting findings from these studies, we aim to make visible how AI has the potential in such settings to either disrupt or support that reasoning in various ways. In view of this, we argue that it is important that design endeavours for xAI in this domain start out from a reasonable understanding of just what that reasoning looks like.

As noted below, it has long been standard practice in CSCW and HCI to use ethnographic studies as a way of examining existing practice and explore the kinds of impact the introduction of technology may have. In dialogue with designers, these provide a way of testing the assumptions present in technology design about the settings in which technologies may be deployed in order to see whether they are well-grounded in an understanding of the social organisation of those settings. This illustrative and instructive approach forms the backdrop to the studies we report here. We will specifically be making use here of ethnographic data from three studies relating to the diagnosis and treatment of breast cancer: the expert practices involved in reading mammograms; the cross-disciplinary reasoning that is brought to bear upon the treatment implications of expert recommendations in what are called Multi-disciplinary Team (MDT) meetings in a breast cancer unit; and a trial of a decision support system for mammography screening. Ethical approval was obtained for each study and participants gave their informed consent.

It should be noted that the specific rationales for undertaking each of the studies differ from one another in certain ways. In the first study, the focus was upon how radiologists working in breast screening decided on whether to refer a patient for further investigation (Hartswood et al. 2002a). The second study was undertaken to inform the development of prognostic tools for use in MDT meetings, illustrating how not only the diagnostic process is being opened up as a space within which to deploy AI but it is also envisaged as a resource to underpin prognosis. Diagnosis and prognosis come together in the context of MDT meetings that are held pre- and post-operatively. The third study was interested in the impact of the introduction the decision support system for mammography screening and how expert reasoning and machine-based recommendations rub up against one another when they are obliged to work together (Hartswood et al. 2003, Slack et al. 2010).

There is a process involved in the diagnosis and treatment of breast cancer and we have sought to selectively illustrate here the organisational accountability and reasoning associated with this. The first example in section 3.3 relates to how radiologists examine images for potentially cancerous lesions. In section 3.4, we see how the findings of various healthcare professionals are collectively brought to bear in MDT meetings to decide on how to proceed with the treatment of any identified cancers. Thus, there is a strong relationship between the work of and organisational accountability of radiologists described in section 3.3 and the work carried out in MDT meetings, where specific expert accounts are brought to bear. MDT meetings are effectively where diagnostic outcomes produced by individual experts (including any outcomes informed by technical apparatus, such as, potentially, AI) are parleyed into a treatment plan and, simultaneously, made accountable to all relevant organisational stakeholders. This is the kind of work described in section 3.4. Finally, in section 3.5 we see how radiologists attempted to make a prototype CDSS accountable for its behaviour. We use materials from these studies to understand how these different bodies of practice are accomplished and what organisational accountability looks like in each situation. This gives insight into the settings within which AI-based accounts will also have to find a home. We would argue that, while technologies may change, the basic processes and reasoning about the organisationally accountable characteristics of moving from diagnosis to treatment, do not.

We are aware that the significance of these studies for trustworthy AI might be challenged on the grounds that they are not recent or that they might not be relevant to AI technologies or healthcare settings and practices today. First, to build upon the preceding remarks, the role played by accountability in establishing and maintaining trust in organisational work and the ways in which it is reasoned about as a feature of that work remains a preoccupation in healthcare. Second, as a socio-material practice, how accountability is performed is shaped by both the technologies and settings in which accounts are presented, interrogated and shared. However, to assume that AI, by virtue of its presentation will bypass the socio-technical concerns we elaborate and be 'trusted' by virtue of its technical features alone, is to assume that the sociality of such settings will be held subservient to such considerations. On the contrary, the materials we present below make it manifest that the opposite is the case: all organisational settings are invested with an order that is first and foremost moral and social and within which any kind of technology must find its home. As observed many years ago by Sacks regarding the introduction of telephones into domestic environments, technologies are "made at home in the world that has whatever organisation it already has" (Sacks, 1992, vol. 2: 548-9). So, while the specificities and possibilities of technologies may change, the fact that they will be subjected to the kinds of moral reasoning we describe does not.

**3.2 Methodology**

The research in the three empirical cases studies was carried out through the deployment of an observational method known as 'ethnomethodologically informed ethnography' (EIE) (Button et al. 2015). Following the 'turn to the social' (Anderson 1994, Button and Harper 1995, Jirotka et al. 2005, Randall et al. 2007), this approach to understanding the activities and processes connected to everyday work has become especially prominent in both computer-supported cooperative work (CSCW) (Luff et al. 2000, Randall et al. 2005) and human-computer interaction (HCI). The aim of EIE is to identify and describe the particular activities in any workplace setting, to consider the myriad ways in which work is accomplished, moment by moment, irrespective of, and 'indifferent' to, any pre-existing theoretical or organizational stance on how work 'should' be done. As Randall (2018) puts it, "Roughly speaking, this entails a commitment to the point of view of the actor; some kind of preference for study of the way actors order their activities; an interest in the skills, competencies and 'artfulness' that actors bring to their efforts, and an interest in the use of artefacts."

The general warrant for EIE is that of 'faithfulness to the phenomena' – the thorough description of the situated organisation of activity in all its real-world detail. It sets out, as Garfinkel puts it; "to treat practical activities, practical circumstances, and practical sociological reasonings as topics of empirical study, and by paying to the most commonplace activities of daily life the attention usually accorded extraordinary events, seeks to learn about them as phenomena in their own right" (Garfinkel 1967). In our analysis, we focus on what we can learn from the real-world, real-time competences and practices through which members of the setting organize their interactions. This involves fine-grained, moment by moment, analysis of everyday situated practices and interactions in order to explicate people's 'ethno-methods' – the practical, situated exercise of common-sense, whereby activities are made to be seen accountable, organized and recognisable. In everyday working life people just 'get on' with things; and its exactly that 'getting on' that we describe and analyse. The emphasis throughout is on documenting the actual 'doing' of work: how mammograms are 'read'; how MDT meetings proceed and make decisions; that is, how work is done in actual practice. It is this detailing of social interactions that ultimately makes the findings relevant and important to design activities (Dourish 2006).

**3.3 Reading mammograms in the UK Breast Screening Programme**

Our first case study is of the UK breast screening programme, where mammograms were 'read' by two radiologists and the recall/no recall decision was made on the basis of these two independent assessments (Hartswood et al. 2002a). In Figure 1 a radiologist is examining mammograms (2 views: 'oblique' and 'CC'). As accounts, mammograms may seem of limited value, but radiologists are able to work up a professionally relevant explanation of what they can see in the mammograms by, e.g.: (a) comparing features across the 2 views; (b) using a magnifying glass; and (c) measuring features using their hands (Slack et al. 2010).

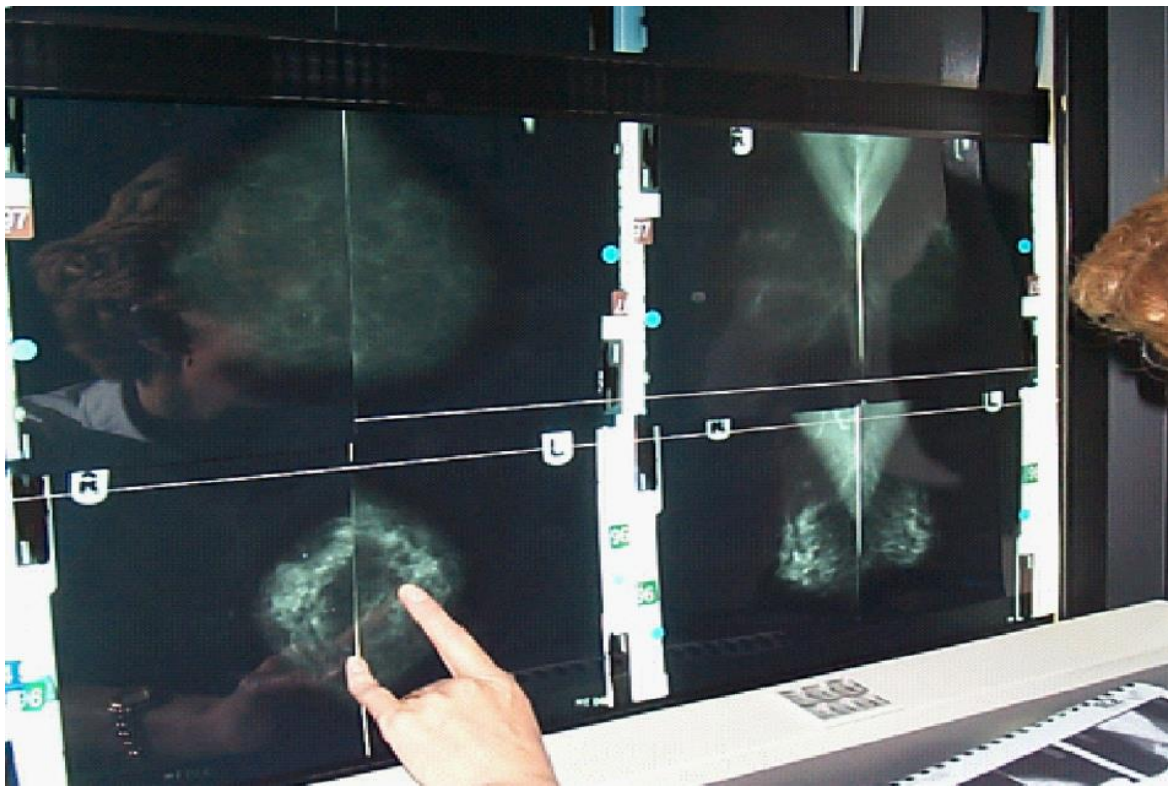

Figure 1: Radiologist examining a mammogram, showing use of fingers to measure a feature.

In reading mammograms, radiologists are required to exercise a combination of practical reasoning, knowledge and skill, as they translate the features visible in the mammogram, their knowledge of its underlying physics, breast architecture and circumstances of the patient into an appropriate organisational account. This requires a range of professional skills to find or rather 'unpick' or uncover what may be faint features in the complex visual environment of the mammogram: interpretative skills to classify them appropriately as being 'benign' or 'suspicious'; and explanatory skills to make their procedures and decisions professionally and organisationally accountable. Types of features that are indicators of malignancy include: micro-calcification clusters (small deposits of calcium visible as tiny bright specks); ill-defined lesions (areas of radiographically dense tissue appearing as a bright patch that might indicate a developing tumour); stellate or spiculated lesions (visible as a radiating structure with ill-defined borders). 'Architectural distortion' may be visible when tissue around the site of a developing tumour contracts; asymmetry between left and right mammograms may be the only visible sign of some features.

Radiologists' 'professional vision' (Goodwin 1994) entails being able to distinguish between what is 'normal' and what is 'abnormal' through an understanding of 'territories of normal appearance' and 'incongruity procedures' (Sacks 1972) and a repertoire of physical manipulations that make such differences visible and accountable. When we look at the work of reading, we observe it as a skilled, reasoned, and above all accountable, practice. For radiologists, diagnosis involves 'undressing' or 'picking apart' features that appear on the mammogram. For example, the important characteristics of calcifications are their size, shape or morphology, number, and distribution. Benign calcifications are usually larger than calcifications associated with malignancy. They are usually coarser, often round with smooth margins and are much more easily seen. But microcalcifications are generally very small and so may be missed; in regions where the background tissue of the breast is dense, it is very difficult to localize the calcifications; moreover, calcifications sometimes have a low contrast to the background. Sometimes, other structures in the breast may mimic micro-calcifications – such as calcified arteries – or artefacts on mammograms (e.g., due to specks of dust or talcum powder) may look like micro-calcifications. These characteristics are revealed through the process of 'undressing'. Such 'undressing' is accompanied by relevant descriptors that invoke a repertoire of descriptions of significance concerning shape, size, density, contour, number of flecks, distribution, orientation, location etc.

Figure 2 shows the screening form used to record comments and decisions made during the process. It consists of several distinct sections that are intended for the use of the different members of the screening centre team. The process begins when the mammograms are taken by the radiographer, who records information about, e.g., information gleaned from the woman about her medical history (e.g., cyst; moles; Pain L breast and arm, GP thinks is muscular), together with details of how the x-ray machine was set up when the mammograms were taken. In summary, these are observations that are expected of a competent radiographer and which a radiographer records to make their actions accountable. Comments are indexically tied to the mammograms through marking the simple schematics. Similarly, the first radiologist to examine the mammograms doesn't just record recall/no recall but adds a comment (e.g., new), which is then available to the second radiologist, who adds a final comment (e.g., BT I think, HRT related). The combination of images and forms provides the means for radiographers and radiologists to make themselves accountable for their actions. As a record of decision-making, these accounts may be re-visited should there be any subsequent questions about the original decision (Hartswood et al. 2002a).

Figure 2: Example of a UK Breast Screening report form.

**3.4 Decision-making in multi-disciplinary team meetings for breast cancer**

Our second study was of multidisciplinary team (MDT) meetings for breast cancer. MDT meetings facilitate various processes and interactions around the detection, diagnosis and treatment of cancer. The study was focused on developing insights for the creation an enhanced version of the Nottingham Prognostic Index (NPI), NPI+, which uses a naïve Bayesian classifier to predict patient outcomes (Soria et al. 2010, Rakha et al. 2014).

To make NPI more directly usable and relevant in MDT meetings, a general question was posed regarding how clinicians and other healthcare professionals make decisions and develop those decisions in their interactions with both one another and with patients. A key aspect of this is how such decisions are made accountable and how they feed into a process of accountability. The study sought to understand what informed these decisions and the implications of that for the design of future prognostic systems and representations. In these meetings 'accountability' refers to not only some sense of responsibility for the production of an image, or the making of a diagnostic or treatment decision; but also the making of an action 'accountable' in the sense of it being made visible and apparent to others as the action it is, and repairing any misconceptions or misunderstandings. Accountability is then woven into the whole process of MDT meetings and not merely occasioned by some particular decision; indeed, the fact that it is a 'process' forms part of its accountability.

In MDT meetings, health professionals display, make use of, and found decisions upon various kinds of information relating to patients, including the NPI score (based on factors such as tumour size, grading, etc.), which are input into a prognostic index formula (Rakha et al. 2014). Our interest, as far as accountability is concerned, lay in understanding the grammatical constitution of accounts in MDT meetings: the kinds of information that gets displayed, when it is displayed, how it gets displayed, the specific situation in which such display is embedded, and the ways in which displays of information get cued, and how changes of display are cued and managed. While each particular MDT meeting has features that are exclusive to the particular setting, the general process will be familiar across a range of settings: there will be some form of 'family resemblance', the most important of which is the strong theme of accountability that runs through the process. As noted in section 3.1, MDT meetings are also where the outcomes of using diagnostic tools, such as those used by radiologists and pathologists, are brought to account. Thus, MDT meetings constitute a key environment to understand with regard to how expert decisions are made more broadly organisationally accountable and opened up for potential further inquiry and explanation.

In the MDT meetings we observed, the clinicians managed the running order by making reference to printed sheets provided in advance by admin staff. The meetings took place at the beginning of each day before patients visited the clinic and the printed sheets gave the order in which the patient visits were scheduled. At the start of each patient discussion, the clinician responsible for that case would provide a short history, giving origins, actions and current status, including any technical information, if known. A radiologist would then continue with the radiology-derived information. Where relevant, radiological imagery on one of the screens at the front of the room would be displayed, as specifics were mentioned. Details from this were typically noted down by the clinicians and nurses. When the radiologist had concluded, it was the turn of the pathologist to deliver a summary of the pathology report based on biopsies or surgically extracted material, including technical figures. While the pathologist was talking, the clinician, nurses and administrative staff all took notes. The pathologist then handed over the printed report to the responsible clinician, who then took further notes from it prior to inserting it in the patient record. After this, a decision regarding next steps was made by the clinician, either as a direct proposal or in discussion with others in the room. This could be short or more protracted, depending on the case. Once a decision for treatment had been made it was noted on the MDT meeting record by the clinician. Admin staff simultaneously noted the same information on their own copies of these records and nurses noted it on their Breast Care Nursing Assessment Forms.

It can be seen from this overview that MDT meetings are shaped by a process that is all about accountability. At each and every stage, people are required to give accounts of or explanations for their decisions. Accountability also attaches to the process itself in the form of who has the right to speak and the need to account for interventions or comments. For example, nurses may provide reference to, clarification of, or enrichment of patient biographical details. Radiologists and pathologists may elaborate on, clarify, and/or enrich technical information when it is mentioned by the clinician, especially where uncertainty about some technical or pathological component is voiced. Similarly, admin staff may add to the nurse commentary or elaborate upon aspects of the record and its interpretation where this is unclear, or errors are manifest. There are global and local grammars to be followed and exceptions to this may matter. Surprising or unusual results can provoke debate. When these kinds of discussions occur, the other clinicians may request additional information from nurses, pathologists, radiologists, admin staff, other clinicians; in short anyone present with specific and relevant knowledge.

Another observation concerns how accountability is afforded by and accomplished through the visibility and sharing of particular forms of document – paper records, slides displayed on screen, etc. A number of items were visible to everybody in the room by virtue of two screens on one wall displaying a range of information. This included: a list of patients under discussion; the radiology patient record containing a summary of the main features of the patient's case and scanned copies of relevant documents; radiology images – CT scans, mammograms – biopsy images, NPI-derived survival curves, individual drawings and sketches; letters from GPs etc.

The displayed record can become quite specifically a talking point within discussion of ways to proceed – especially where specific events or findings in the past are deemed relevant – with it standing directly as an object of ostension, people in the room being able to point to aspects of the record and refer to them with them being visible to all. Within the course of the meeting, items in the record are made use of by specific people, with the organization of the display and use of these informational resources founded upon the ongoing flow of talk, with talk cueing certain elements and then being used to describe and elaborate upon what is seen. Much of the decision-making hinges upon factors such as the size, the category, the type, the margin, whether there's been any vascular invasion, what was found in the lymph nodes, the NPI score and so on. All of these can implicate different patterns of treatment.

Arriving at accountable decisions is something that unfolds dynamically. At the same time, different parts within the process of diagnosis, treatment and care implicate the production of certain kinds of record. Thus, the making of decisions and the recording of decisions can be a hugely interwoven and mutually elaborative affair, rather than it simply being a case of 'facts are presented', 'decisions are made', 'related facts and decisions are recorded'.

So, within the course of MDT meetings, a number of resources are called upon; that is, are called to account. Patient files, for example, are managed by clinicians and have a rich and important relationship with the process of accountability and the related decision-making process – they form part of its stylistic unity (Bittner 1965). They feed into delivery of the patient history and stand as a resource for elaboration. At the same time, they are an accountable auditing device, standing as a repository of decisions made and where one might find a record of past states and actions and whom they relate to. In addition, and as an outcome in part of the meeting, they give access to the current state and projected next actions.

Clinicians enter details from radiologists and pathologists during the MDT meeting and the decisions made as a consequence on MDT meeting record forms. These are headed with an administration section and have sections for completing information as it is provided by the radiologist and pathologist and then a space for recording the decision. The forms cover a wide range of technical details, and some details may be appended or changed as things are checked and verified or elaborated upon. In this sense the forms provide a schematic record of the in-situ elaborations. The decision is entered as the discussion concludes. These forms are always completed in the meeting in the context of each case. This makes the MDT meeting in part about record creation and annotation – not just presentation, display, talk and decision-making.

The responsible clinician is also accountable for documenting what unfolds next: as the radiology report unfolds; as the pathology report unfolds; as elaborations are provided; as a decision is arrived at (or not). This documentary work has to be managed within the course of the interaction, which has implications for how clinicians engage with other resources during the meeting. In terms of the decision-making process these forms become the accountable representation of the decision and (ideally) provide the materials within them, and in relation to the rest of the patient file, for the decision to be accountable and intelligible to other competent members. Additionally, as documents that record the pathology data as it unfolds, they can also stand as a reference point for information during interaction leading towards the decision.

**3.5 WHAT DOES IT MEAN FOR AI TO BE ACCOUNTABLE?**

In this section, we examine how healthcare professionals reacted to the challenges of using an AI system that lacked the capacity to explain its behaviour.

In the past two decades, breast screening has been the site of several attempts to introduce CDSS into the process of reading mammograms. Figure 3 shows a prototype of one such system that was designed to substitute for the first radiologist[1]. At the top are the original mammograms and at the bottom are displays showing prompts, that is marked-up areas in the mammograms that the system's analysis of the mammogram has decided are suspicious. We studied radiologists using this CDSS in order to understand how they made sense of these prompts and how the prompts influenced their decisions (Hartswood et al. 2003, Alberdi et al. 2005, Slack et al. 2010).

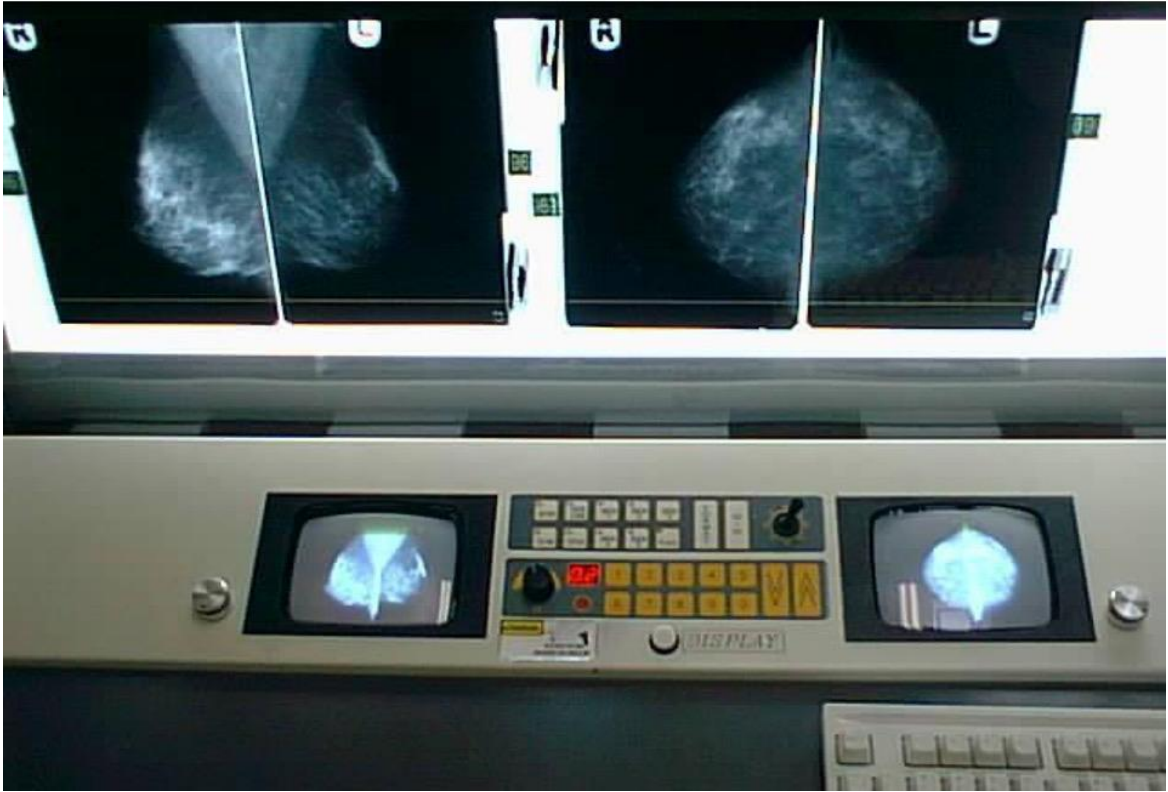

Figure 3: The breast screening decision support system showing the prompt viewer.

The system generated prompts that call attention to particular features on the mammogram but provided no explanation for them, leaving radiologists with the task of providing some form of 'natural' accountability for the features it highlighted. This was especially noticeable when radiologists disagreed with the system because the radiologists then had to explain, i.e., to account, why the prompt should be ignored or, alternatively, acted upon (see Figures 4 and 5) (Hartswood et al. 2003, Slack et al. 2010).

---

[1] Using AI to substitute for human expertise remains a widely used argument for CDSS deployment. See, for example, McKinney et al. (2020).

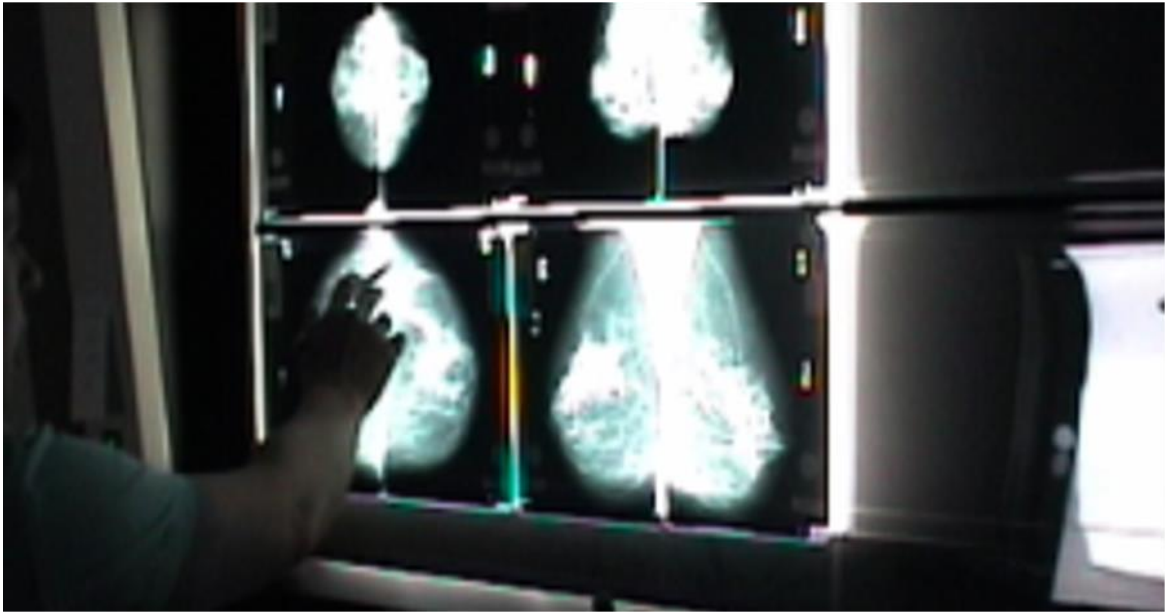

Figure 4: It's got a whole row of markers up here… but they're all on innocent things… nothing to worry about there.

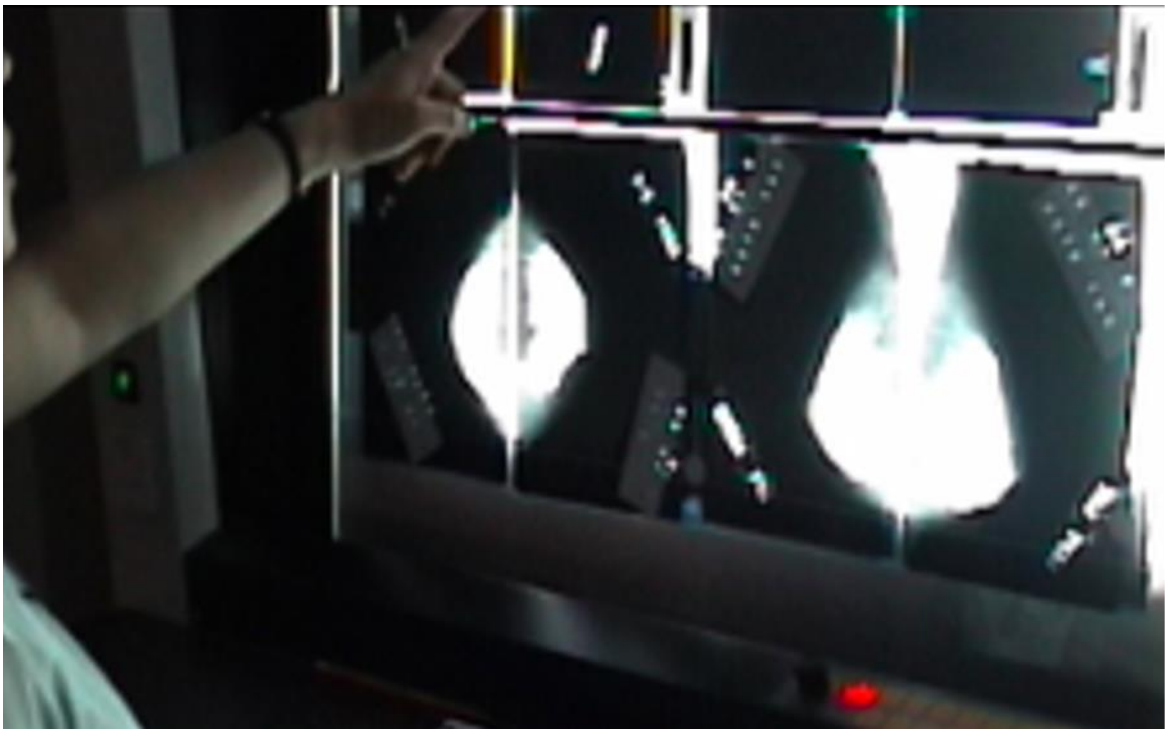

Figure 5: I've seen this before… the computer's marked something that I think is artefact on that side… it's often along the edge of the film.

Partly as a consequence of its inability to provide accounts of its behaviour, this system failed to demonstrate improvements in real-world settings and was never adopted (Alberdi et al. 2005), and this remains a persistent problem for AI systems in healthcare (Yang et al. 2019).

**4 DISCUSSION**

We now examine in more detail the various dimensions of accountability visible in the three case studies in relation to the review of accounts and organisational accountability presented in section 2.

**4.1 The accountable characteristics of reading mammograms**

The report form records radiologists' reading of a mammogram, including the various features uncovered through the process of undressing and the grounds for their recommendation, i.e., 'recall', because their reading indicates that certain features are 'suspicious', or no recall because they consider the features to be 'benign'. Once completed, the report form stands as a formal account of the reading procedure and has a relatively generic status in that it can be consulted by a range of others who are not a party to the situated reading and generation of the account itself. This is, of course, not quite the whole story because the form is open to contestation and modification by the second radiologist. Nonetheless, once the reading process is complete, from the point of view of medical records, the form stands as a formal account of what took place.

Close inspection also reveals that what might stand as a formal account also has elements here of situated accountability. Two things are noteworthy in this regard. First of all, the form works to make visible what would otherwise be opaque to non-competent others in the original situated reading of the mammogram. So, it is a formal account of a situated reading and is negotiable in those terms should some other party inspect the same mammogram. Second, the form is itself formulated by the first radiologist for reading by the second radiologist. In other words, it is recipient designed (Sacks 1992) for someone else with a shared understanding of the work of reading mammograms, not just anybody. So, it is clearly not really generic but rather targeted at other members of the same cohort with a similar competence.

In the case of the breast screening CDSS, however, the study makes clear that there was little 'accounting' going on. For the system, every marked-up entity had the status of suspicious and it was left to the radiologists to arrive at an actual account in situ, though their own record of the reading of the mammogram may then acquire the status of a formal account. This is indicative of some of the issues that trustworthy AI needs to address, because it illustrates the problem when such systems effectively generate just one half of what might be considered an accounting 'pair', i.e., offering an assertion (tantamount to 'suspicious' on the original reporting form) without an explanation to underpin the assertion (the reasoning uncovered through the undressing process articulated through the original form).

As the CDSS was not capable of providing an account for its behaviour (i.e., the presence or absence of prompts), radiologists were left to come up with one based on their accumulating observations, a 'biography' (Slack et al. 2010) of the system's behaviour that is, in itself, a kind of account, one that might stand in equivalence with a formal account in that it is global and generic (e.g., Figure 5: "I've seen this before… the computer's marked something that I think is artefact on that side… it's often along the edge of the film"). To serve as an organisational account, it would need to be shared between – annotated and curated by – radiologists as their experience of its behaviour accumulated with every prompt they saw. However, unlike the ways in which, as we saw, radiologists were able to make a mammogram an accountable object and which drew upon their understanding of its physics, their knowledge of breast architecture and how these inform procedures for 'undressing' a lesion, this system biography was not informed by radiologists' detailed knowledge of how the system worked but by accumulated observations of its behaviour, and so was likely to be partial and inaccurate. There is, therefore, also a need for a situated account, that is one that is tailored to the particular behaviour being observed at any moment.

Again, all of the work of arriving at a situated account falls upon the radiologist because it is absent from the account generated by the system. So, radiologists cannot just take the system's behaviour at face value but rather have to account explicitly for either disregarding or acting upon it. In the case of contestable prompts, there is a need to account for their disagreement and to offer up an alternative account in its place, i.e., conduct the work of repair. So, it is clearly not the case that the marked-up mammogram might be said to constitute something like the first radiologist's report. In double reading, the first radiologist's report is designed to make visible solely relevant features based upon a common understanding of what a relevant feature might look like. The marked-up mammogram has none of this implicit understanding, so all of the work is left to the remaining radiologist. This shows the economy of the recipient design based upon the first undressing and also reveals that, as things stand, the reading of a marked-up mammogram might be said to generate additional work. This is clearly not ideal for an AI system designed to offer support and reveals starkly and why this system could not be used to substitute for the first reader.

The false prompt rate of the CDSS in this study was high (Champness et al. 2005) and it might be argued that accountability would not be necessary for a better performing system. Three observations are worth mentioning in response. First, the performance of an AI system may vary: for different classes (in breast screening, these classes may reflect, e.g., type of lesion: micro-calcifications, masses, spiculated lesions, etc.); with different data (Oakenden-Rayner et al. 2020); in different contexts (Sanneman & Sha 2022); and may subject to 'drift' (i.e., change over time), all of which will be difficult to detect without access to accounts of its behaviour. Second, the ways that radiologists are seen to orient to being accountable for their decisions shows radiologists understand that evidence of their trustworthiness has to be continually reproduced in and through the course of their work (Clarke et al. 2006). Third, and relatedly, an AI system whose performance is at a similar level to that of a radiologist (e.g., McKinney et al. 2020) would still be required to provide accounts of its behaviour, because, as we argued above, these are a material resource for the account clinicians will be expected to provide for their decisions.

Beyond this, there are clearly ways in which natural accountability comes into play in the work of radiologists in this case study. A great many of the procedures and descriptors used in the course of doing the work are assumed to be appropriate, without any explicit account being given for their use. This includes both the matters of focus and report when 'undressing' a mammogram and the order within which that 'undressing' is pursued and reported. This attests to an oriented-to, intersubjective grasp of what is relevant and appropriate on the part of other organisational members who might engage with the work. There are ways in which both the procedures of undressing a mammogram and the procedures for reporting what is revealed are also possessed of various features of grammatical accountability, as discussed in section 2: these things cannot be done in just any order. The choice of terms to describe the features found attends to what might appropriately co-occur and do the job of mapping out a topology of suspicion that might be expected to be reasoned about in the same way by another party with similar interests. Within the cohort of radiologists, there is also a set of operational MCDs, whereby certain terms, such as 'lesion' and 'margin' might be seen to go together. All of this is never rendered explicit, but it constitutes a 'backdrop' upon which effective reasoning and diagnosis might proceed. This backdrop, or set of background expectations, forms a central part of what a presumptively adequate, trustworthy AI would need to be able to grasp to be able to measure up to what experts take to be a demonstrably capable professional in their domain. As we noted in Section 2, the capacity to draw upon the same background expectations is also a central component of trust.

In fact, there are both formal and taken-for-granted aspects of trust visible in the work of the radiologists. One of the key formal mechanisms for ensuring the 'trustability' of accounts is the need for two independent assessments. Implicit in this are two considerations: i) if two separate experts provide similar accounts for what is visible, the accounts can be 'trusted'; ii) if one radiologist produces a contestable account, there is an opportunity for the contested account to be itself called to account, negotiated and, if necessary, repaired. Both of these matters are also further formalised and, in a sense, made more explicit by the use of the reporting form. All of this, in turn, is also bound up with both the association between trust and background expectations (the intersubjective point of view) and the points made by Garfinkel regarding the use of formal records and organisational accountability. There are multiple ways in which the elements entered into the report trade upon background expectations of what specific terms and their co-constitution might amount to when delivered in a formal report. Clearly, also, the recipient design of these elements for another radiologist involves knowing things like the organization (i.e., the NHS breast screening programme), its operating procedures, and what, through the elliptic terms provided, will evidence the adequacy of the radiologist's actions for others in the same organisation.

This can also be seen to resonate with Bittner's notion of a gambit of compliance (1965), in that it makes manifest an understanding of the organisation and how it works. For the other radiologist in receipt of the report, the report is only going to be sufficient for the work in hand if they have knowledge of who the report is referring to, who contributed to the report (i.e., the first radiologist), and, potentially, their mutual history with the other radiologist in the same organisation, with known operating procedures. Most importantly, as the report is, in Garfinkel's terms, a 'therapeutic contract' (Garfinkel 1967), the other radiologist knows the procedures for being able to read it in an appropriate fashion. A curiosity in this case is that, unlike the records Garfinkel and Bittner examined, these reports are open to being revisited by other parties as a matter of open policy, if there is any question about the decision, which means that there is also a sense in which they have to work as actuarial records as well. This, of course, begs a question as to who might be a competent 'reader' of the record under such circumstances. This actuarial component is provided for by the structure of the forms, the fields given, etc., but, clearly, this exposes the radiologists to being potentially called to account for, not only their decision, but the adequacy of their form-filling practices. There are ways in which this is covered by Garinkel's exposition of organisational accountability. Knowing what is actuarially adequate is a part of knowing an organisation and its procedures. Here, however, the 'organisation' to which one might be deemed accountable is somewhat large and diffuse, including one's professional colleagues, the clinic for which they work, the NHS breast screening programme and its regional and national entities, medical professional associations, legal authorities of various kinds, and even the recipients of care and their representatives. All of this is to say that, in the case of safety critical services, there is a mixture of the therapeutic and actuarial considerations in how records are constituted that modifies some of the character of organisational accountability. As we note below, this gives a particular twist to how matters of accountability and trust might play out when incorporating AI systems within the diagnostic and reporting practices of radiologists and within healthcare more broadly.

**4.2 The accountable characteristics of MDT meetings**

MDT meetings are where formal and situated accountability come together. Consultants, radiologists, pathologists, nurses and administrators work together to provide a situated sense of what disparate formal accounts amount to. They do this to construct the biography of both a patient and their illness and the likely ways in which that biography will continue to unfold. We have seen how MDT meetings are also replete with not just presumptively adequate naturally accountable action, but also callings to account, with specific understandings and their implications being opened up to question. Natural accountability is especially evident in how the taken-for-granted structure, order and pursuit of the process is achieved and the nature of the practices on display. Background expectations, and ways in which the common intersubjective grounds of understanding a person's illness and how to proceed are established, provide for the trustability of each other's competences and, critically, trust in the process and ways in which the biography of the illness has been managed and will continue to be managed.

In terms of organisational accountability, compliance with the right way of doing things is evident in MDT meetings, both in participants' adherence to the procedures whereby the meeting can progress (using the printed sheet regarding the 'order of the day', etc.) and to the documentation of the meeting and the decisions made. There is a sense in which the whole thing is about properly managing the therapeutic contract while ensuring the actuarial contract is validated and kept up to date.

With regard to the contract itself, knowing who the record refers to is procedurally established from the outset, but is also subject to elaboration and update. Each party has a certain interest in how the patient's biography is constructed. Although these interests are not commensurate, they are mutually elaborative and, for the larger part, understood to be complementary. One of the strong organisational affordances of MDT meetings is that they make especially available who is contributing to the record because much of the record is either actively being presented by the person who prepared it or is being cooperatively constituted then and there in front of one another. This makes the accountability of the record producer and the practices they adopted for preparation of records directly available.

The whole MDT meeting process, with the parties present, the timing of the meeting, the resources made available, the layout of the room and where people sit within it, its running order and the records produced, testifies to a shared and taken-for-granted knowledge of the organization and its operating procedures amongst everyone present. Also, MDT meetings are largely populated by the same personnel: despite their diversity of organisational roles, all parties come to have a mutual history that is itself a feature of the ways in which the resources made available are treated as either naturally accountable or open to being called to account. A wide variety of records and information derived from them are put on display. In this way, through the presentation of the records and their situated accountability, members of MDTs come to acquire knowledge of the procedures for reading the various records. This is not to say that nurses can read a mammogram as well as a radiologist, but they are instructed through the course of the meetings in what a competent reading of the record looks like, such that they would recognise anomalous practices and be able to call them to account. As stressed in the description of MDT meetings in the case study, they are all about accountability. They provide for a recognition of the adequacy of each other's actions for everybody else in the local organisation, regardless of role, and, most importantly, they provide for recognition of the adequacy of the decisions made, they provide an opportunity for those decisions to be called to account by competent others and, ultimately, through the procedural mechanisms they provide (technical and interactional), they make everyone in the room potentially accountable for the decisions made.

In relation to Bittner's (1965) view of organisational accountability, everyone in the room has an understanding of what organisational compliance looks like, so there is a sense in which every account put forward is a gambit of compliance because it contains within it the means for others present to recognise its organisational compliance (Zimmerman 1971). As noted in the case study, MDT meetings actively provide a mechanism whereby the stylistic unity of the organisation can be accomplished. They also provide, by the same token, a source of corroborative reference (Bittner 1965), because every taken-for-granted moment of adequacy within them provides everyone present with the sense that 'yes, we do all understand this organisation and its goals in much the same way, even though we have disparate roles to play within it'. Indeed, it is hard to think of a better articulation of Bittner's notion of corroborative reference than this.

Another way in which MDT meetings reflect existing understandings of organisational accountability within the literature is that they are manifestly a mechanism for enabling prioritisation: prioritisation of patients and prioritisation of the handling of those patients and the treatment they receive. They also provide a mechanism for making decisions that are visibly in line with organisational policy. Indeed, one of the grounds upon which decisions actively do get called to account in MDT meetings is the extent to which they adhere to understood ways of dealing with this particular kind of case.

In relation to grammatical accountability, MDT meetings are replete with a situated grammar that is almost never made explicit, yet upon which all other matters may be seen to turn. The whole meeting has a strong spatial organisation that positions certain parties in certain places in the room and certain kinds of display in certain evidently visible locations for those who have to see them and refer to them. Even the spatial organisation of the patient files has an accountable order that drives the sequential order of the meeting and that is presumptively adequate to all involved. The temporal order of the meeting is, of course, a playing out of this spatial layout, but it is also organised around a clear understanding of turns and just when different parties have a right to speak. Even the constitution of radiology and pathology reports has an internal consistency of terms that makes them manifestly what they are. The contents of these two different reports, or nurses' reports or even consultants' reports are not simply interchangeable and, were they to be changed in such a way, the competence of the person compiling the report could well be called to direct account. So, in the end, the description of a patient and the specific features made relevant within that description, are, in a strong sense, a mutual accomplishment of the MDT meeting. Descriptions are either ratified, negotiated or revised on the basis of how the MDT meeting unfolds. The grammatical constitution of the meeting itself is also a product of how the resources within it are mutually co-assembled and how all parties present are complicit in seeing that co-assembly as natural, appropriate, probative and productive of trust.

**4.3 Understanding accountability as a constraint and a resource for AI**

The preceding analysis of how organisational accountability is achieved illustrates the challenges that may be thrown up by introducing AI systems into specific healthcare settings and work practices. We will now consider the implications that follow from this analysis for trustworthy AI and which we argue are not being taken into account in current research. In particular, our findings suggest that trustworthy AI is likely to require a range of different forms of accounts, perhaps assembled in various ways and recipient designed to match the needs of particular on users, setting(s), timings and the circumstances in which these accounts are deployed. Our analysis draws on observations of healthcare work in real-world settings seen through the lens of organisational accountability. This contrasts with, for example, the work of Cai et al. (2019a), who conducted user-based evaluations of CDSS diagnostic utility and Ehsan et al. (2021), who used a scenario-based methodology in a series of interviews with 'stakeholders in AI-mediated decision-making domains', including healthcare. It should be noted that none of the participants in the latter study were healthcare professionals.

We wish also to stress that it is not our aim to decide on the suitability in healthcare of the various techniques for xAI now available: such judgments are necessarily specific to the application and its setting. As Ehsan et al. (2021) remark, AI systems and explanations are socially-situated, which requires incorporating the social aspects when conceptualising xAI. In what follows, we summarise lessons from our case studies on how grammars of accountability make distinct demands on accounts and what, in a more general way, these imply for requirements for xAI and trustworthy AI.

Earlier, we remarked on how, on the one hand, current thinking on xAI in relation to trustworthy AI distinguishes between *global* and *local* accounts and, on the other, the literature on how people do being accountable distinguishes between *formal*, *situated* and *natural* accountability in the context of organisational work. It seems to us that a step forward in understanding requirements for xAI and trustworthy AI would be to: a) examine the ways in which these two distinct ways of thinking about accounts may be related to one another; and b) their grammatical constitution within the healthcare pathway, that is, their spatial, temporal and coherence concerns. With this in mind, we present a series of scenarios for AI accountability, together with some suggestions on how this might be satisfied in each one.

First, we argue there is a parallel between global xAI and formal accounts. The final stage in the development of an AI system is validation, where performance is measured to establish its accuracy (Liu et al. 2019). For example, where a system is to be used to classify features in a mammogram as either 'normal' or 'suspicious', the validation step will establish: a) how many cases were classified correctly (i.e., true positives and true negatives); and b) how many were misclassified (i.e., false positives and false negatives). These four measures can be combined to calculate the system's accuracy (Ferri et al. 2009) and hence provide a measure of its trustworthiness. The mammography CDSS case study reveals that one of the radiologists' ways reasoned about its trustworthiness was the region of the mammogram within which prompt were located. This suggests that, for AI applied to medical imaging, information on the system's accuracy for different types of features in relation to distinct regions and 'geographies of suspicion' could be an important element of a global account.

Second, given that a global account is designed to explain an AI system's behaviour as a whole, it may not provide sufficient evidence for assessing a system's trustworthiness in a particular case. For this, local xAI, that is, a situated account specific to the current case will be required. Providing local xAI is generally easier than global xAI and various local xAI techniques have been developed for AI systems in healthcare (Poceviciute et al. 2020, Zhang et al. 2022). Saliency maps, a visual display highlighting features that contribute to a prediction (e.g., heat maps), are a natural fit with AI systems that make predictions from image data and are also intuitive (Guidotti et al. 2018, Machlev et al. 2022, Graham et al. 2022). Couture et al. (2018) tested a saliency map technique in an AI system for detecting breast cancer tissue. Counterfactual xAI is based on examples where minimal changes in the data lead to a different prediction and is said to provide an intuitive way to interact with the AI system (Henderson et al. 2020). Wu et al. (2021) employed counterfactual xAI in a system for lymphedema diagnosis. As xAI techniques differ in the ways in which they provide explanations, they have different strengths and weaknesses (Saarela & Geogieva 2022). A truly situated account would be one that matched the characteristics and accountability requirements of a specific case, so this suggests that different kinds of accounts could be provided from which users could select and employ sequentially and/or in combination. This could also be a solution to the problem of providing progressively richer and recipient designed explanations of behaviour (Button & Dourish 1996) and go some way to providing a degree of natural accountability in AI systems. For AI systems designed to assist in the diagnosis of medical images, a more general recommendation would be that local account design should be informed by the practices that healthcare professionals use to interpret images, i.e., clinically meaningful features (Graham et al. 2022, Nix et al. 2022) and how underlying physical structures manifest themselves. In mammography, this would suggest that local accounts also be sensitive to the 'geographies of suspicion', but also how radiologists' make use of their understanding of the physics of mammograms and knowledge of breast architecture.

Third, we saw in the CDSS case study how radiologists were able to draw on their memory of previous cases – the 'system biography' – in order to help them make sense of its behaviour. There are obvious problems, of course, in clinicians having to rely solely on their memory for previous cases. In a study of prostate cancer diagnosis from biopsy images, Cai et al. (2019b) observed that pathologists may look online for images from similar cases. This led them to develop an AI-based image retrieval tool to enable clinicians to find similar examples from a dataset of diagnosed cases that then can be compared with a new case. This approach could be also applied to a dataset of cases previously classified by an AI system (including training examples). As an aggregated and accumulating record of an AI system's behaviour (Ehsan et al. 2021) and the 'ground truth' of each case as it became available, this dataset would provide a more reliable system biography. This would help meet the need for an up-to-date, global account of AI performance (Liao et al. 2020) but in a form that is distinct from how global accounts have so far been conceived (Das & Rad 2020, Machlev et al. 2022). The inclusion in the system biography of users' case-by-case interactions with the system over the whole of the healthcare care pathway would capture what and how local accounts were used in the making of decisions, and how users accounted for their decisions in the context of the system accounts available, and thus could form the basis for an actuarial record of the decision-making process. Further, by making the system biography queryable on user-selected parameters, previous cases that match a new case could be used as evidence for helping decide on the trustworthiness of the AI system's classification of the new case. For example, if no matches are found, this could suggest that the new case is an outlier and thus the AI system's classification of it may be less reliable (Nix et al. 2022). The inclusion of users' interactions and decisions could also provide a valuable collective learning resource: how colleagues used one or more local accounts when making decisions and how this compares with similar cases may help sustain professional vision as users learn to recognise and adapt to an AI system's strengths and weaknesses (Procter et al. 2022). However, making it easy for diverse types of users to interrogate this system biography, and to do so in different settings, is likely to present significant challenges for data visualisation and HCI design. One setting of particular interest in this regard is the MDT meeting.

Fourth, MDT meetings stand as the point where the outcomes of earlier stages in a patient's journey through the healthcare pathway are assembled into a coherent record that informs what happens next. Indeed, it is these characteristics that have led Yang et al. (2019) to choose MDT meetings as the best site for an AI system to support artificial heart implant decisions. Hence, the MDT meeting's role as the nodal point of organisational accountability, where all the evidence available for each case is discussed in reaching a decision arguably makes it the most significant site in terms of the grammars of AI system accountability, and their spatial, temporal and coherence characteristics. Each use setting has an existing social organisation, so different kinds of accounts will need to fit within that organisation if they are to work effectively and accountably. MDT meetings provide the opportunity not only to identify the form of system accounts, but also how they are assembled as reasoned outcomes, and how they are drawn upon as a source of lessons from each case that can be written into both patient and AI system biographies. It is important then to reflect upon what the grammars of accountability observed in MDT meetings might mean for: the design and choice of global and local xAI techniques as AI system accounts; how a system biography would be made accessible as a both a global and local, situated account to those present; and how its constituent parts would be displayed, annotated and curated in context. As with other kinds of evidence, those present in the meeting will need to be able to understand what a competent reading of the system biography looks like. This raises the question of whether, in MDT meetings, it would fall on the clinician presenting of the case to recipient design their presentation of the system's accounts to meet the needs of radiologists, nurses, etc., or whether this would be factored into how the accounts are put together and what this would represent as data visualisation and HCI design challenges.

Fifth, there are issues raised by the introduction of new technologies into the MDT meeting process. The literature on CDSS adoption demonstrates how challenges of adapting them to the needs of specific application contexts have hindered progress and these will have to be addressed by the new generation of CDSS. In the case of MDT meetings, these challenges would include the possible impact of new informational resources in the form of the various AI system accounts on the existing flow of work. It is also necessary to consider whether they would require significant changes to how MDT meetings are currently constituted. Making AI systems accountable will need to reflect context-specific issues for both forms and assemblies of accounts, for grammars of accountability, and thus for accomplishing the stylistic unity of the process.

Sixth, a healthcare professional's decision-making performance can change over time and for this reason may be subject to monitoring and periodic auditing. For example, radiologists working in the NHS breast screening programme are subject to a range of monitoring and auditing procedures (Cohen et al. 2018). Data and concept 'drift' mean that an AI system's performance may also change over time (Davis et al. 2017a, 2017b, Health 2022), raising the need for monitoring and auditing procedures and tools to detect changes that might put patient safety at risk (Ackerman et al. 2020, Henne et al. 2020, Nix et al. 2022). Providing support for the monitoring and auditing of AI systems (Davis et al. 2019, Liu et al. 2022) would therefore be another scenario to be taken into consideration in the design of the system biography described above. Enabling the interrogation of the system's biography, e.g., for evidence of significant changes in performance, will bring additional data visualisation and HCI design challenges.

Seventh, in many domains, there is a body of competence and practice encompassed within professional vision that is drawn upon to provide situated accounts of various kinds. As the materials in the mammography case study make evident, this competence is observable and, in principle at least, learnable. When it comes to natural accountability and everything that is taken for granted, it is unlikely that any AI system could occupy the position of the intersubjective 'other' in what is taken for granted about an interactant's understanding of the world. Nonetheless, and again as the case studies make visible, there is no reason why some of what is taken for granted cannot be captured through close observation. It is also possible to cultivate expressions of best practice in the domain and to then explore what bringing that about looks like in the actual accomplishment of the work. In this way, although on the one hand, natural accountability may seem to be a constraint upon what it is possible for AI systems to achieve, it is also possible for the grounds of natural accountability in a domain to be articulated as a resource for AI systems to draw upon.

Finally, the common thread between these different scenarios and what it means for AI systems accountability is that their introduction into a setting will shape the practices through which members make themselves organisationally accountable and this is of critical importance. What we want to make clear then is that developing resources for AI system accountability is not a trivial undertaking. Especially in a safety critical context such as healthcare, these resources for accountability, the accountability practices they support and the auditing regimes that ensure they, and the AI systems whose performance they document, remain fit for purpose will need time to evolve. They will also need to be guided by the accumulation and sharing of experience, identified best practices and professional training. This process will, of course, need to be overseen by bodies representing healthcare professionals and by regulatory bodies (ALI 2021). Consultations with patient groups will also be important for maintaining confidence in the safety of AI-enabled healthcare (Richardson et al. 2021, Nix et al. 2022). These, then represent additional scenarios and HCI design challenges for the delivery of trustworthy AI systems.

## 5 CONCLUDING REMARKS AND FUTURE WORK

> "A medical diagnosis system needs to be transparent, understandable, and explainable to gain the trust of physicians, regulators as well as the patient. Explainability is the key to safe, ethical, fair, and trustable use of AI and a key enabler for its deployment in the real world." (Singh et al. 2020)

In this paper we have sought to explore what trustworthy AI means by reflecting on the nature and the roles of accounts and accountability as practiced within organisational settings and illustrating this with case studies drawn from our earlier work in clinical decision-making. It is true that there have been technical changes in healthcare since these case studies were conducted. For example, mammography has moved from film to digital imaging and record keeping in healthcare has moved progressively from paper to digital. As a socio-material practice, the adoption of digital technologies will, of course, shape how accountability is now performed in these settings. However, our findings rest on the self-evident fact that the role played by accountability in timely, safe and dependable healthcare work remains a constant preoccupation.

We have stressed that it is natural accountability that ultimately underpins trust between people, and it is hard to imagine a more effective foundation for trustworthy AI. What we are arguing for, therefore, is resources for accountability that "… can be unremarkably embedded into routines and augment action." (Tolmie et al. 2002). Clearly, delivering trustworthy AI in this way, 'in the wild', will not be a trivial task for HCI design. Trust is a relatively well-developed topic in the HCI and HCAI literature, highlighting some of the difficulties of trust surrounding work activities (Shneiderman 2020, Markus et al. 2021) and suggesting how empirical fieldwork findings might filter down into technical design recommendations. (Schneiderman, for example, outlines 15 recommendations to create reliable, safe, and trustworthy HCAI.) Trust is a subjective assessment of reliability, dependability and other important features of both interactions and technologies, and the case studies reported here explore trust as it relates to individuals, processes, systems, settings and 'trusted data'. They also highlight the idea of trust as both 'constitutive' of, and resultant from, forms of interaction and organisational work, thereby engaging with a significant body of literature from a range of disciplines (Luhmann 2018, Watson 2009). The case studies identify the important role of trust in organisational work, how people 'perform' trust, how it is instantiated in various documents, technical artefacts and procedures and how it enters into everyday work. In contrast to a number of theoretical approaches to trust, the case studies examine some practical instances, thereby developing the concept as an explicit design goal for both technological systems and organizations in order to enable the HCI notion of "trustworthy by design." (Knowles et al. 2014)

There are no simple, technical 'implications for design' to be offered here. As Dourish (2006) argues, this misconstrues the nature of our ethnographic approach, which is a lens for understanding the mundane details of a social setting and illuminating the relationship between various technologies and practice. Ethnographic studies thereby afford various forms of learning about accountability, explainability and the trust they encourage, and the "actors who collectively create the circumstances, contexts, and consequences of technology use." (Ibid), learnings that might eventually work themselves into more technical design recommendations.

In 'The Mechanics of Trust: A Framework for Research and Design,' Riegelsberger et al. (2005) argue that: "… trust and the conditions that affect it must become a core concern of systems development. The role of systems designers and researchers is thus not one of solely increasing the functionality and usability of the systems that are used to transact or communicate, but to design them in such a way that they support trustworthy action and — based on that — well-placed trust." And when it comes to trustworthy AI, as Chatila et al. (2021) suggest, in a comment that seems especially relevant to our examples from the mammography fieldwork, "Learning techniques for processing data to predict outcomes and to make decisions are opaque, prone to bias and may produce wrong answers… Properties such as transparency, verifiability, explainability, security, technical robustness and safety, are key."

Our findings emphasise that an AI system should be capable of being accountable at every stage of the healthcare pathway. They also reveal the MDT meeting's role as the key site for delivering organisational accountability. AI system accounts therefore need to be recipient designed, matching the needs of different users, the contexts in which they are deployed and how they may eventually become part of a larger assembly of information. These factors have important implications for the choice of global and local xAI techniques and for HCI design, which is the level at which much xAI research is currently focused. Further, we argue that there is necessarily a relationship between local, situated and global, formal accounts, with the former potentially being constituent elements of the latter as a 'living biography' of an AI system's behaviour. This system biography would be annotated, updated and curated on a case-by-case basis, creating a record of users' accumulating interactions with and experience of the system, there to be interrogated for a local, situated account as and when required but also to serve as a global, formal account and actuarial record. Matching this complex and expanding dataset so that it is usable at any stage of the healthcare pathway will be a challenge for data visualisation and HCI design, and perhaps especially in the context of MDT meetings. Further investigation of the value of the system biography concept and how to address these design challenges is a key part of our future work plans. There is also the question of how AI system accounts might be linked to the patient record. Policies and practices for the curation of the system biography, that is for adding local, situated and annotated accounts, and capturing how users are interact with them, will need to be designed and design choices will likely be subject to organisational governance procedures and protocols for monitoring and auditing system performance.

The case studies presented in this paper involved significant ethnographic effort and this is an investment that will need to continue to generate the kinds of insights and resources necessary for trustworthy AI – in healthcare or in any other setting – to be a realisable prospect. We have already begun to expand upon the studies reported in this paper by conducting fieldwork studies of pathologists in relation to cancer care in order to obtain a more complete picture of the decision-making work implicated by the treatment of cancer and the places in which future AI systems are most likely to intercept with existing practice (Procter et al. 2022). We also plan to extend our fieldwork to explore requirements for trustworthy AI in a range of other application settings. The understanding gained will facilitate recognition and understanding of 'success' and 'failure' in the deployment and utilisation of AI systems and will be used to inform recommendations for the design of accounts and how to make them available them so that people may interact with them as easily and effectively as possible (Victorelli et al. 2020). We then aim to use the findings to explore the design of accounts with their prospective users.

Ethnographic fieldwork can address what trustworthy AI means within any particular setting, however, there are significant benefits in terms of time and cost savings if their findings are re-usable, at least at the application domain level and preferably at the application setting level. To address this, we propose to develop a toolkit of methods for articulating trustworthy AI user requirements, focusing on linking ethnographic findings to design issues, including identification of design patterns (Martin and Sommerville 2004, Martin et al. 2006) for trustworthy AI design. For example, one particular design pattern might aim to capture the common features of MDT meetings, another the reading of medical images such as mammograms.

We have stressed how the delivery of trustworthy AI, as with any IT system, is a socio-technical problem. Hence, xAI and accounts more generally need to be seen as elements of a 'trust architecture', a configuration of technical and human infrastructure designed to support and sustain trust in AI systems. Elements of the former will include xAI techniques, system biographies and data infrastructures to support performance validation and auditing. Elements of the latter will include governance procedures and protocols, which may require significant adaptations by healthcare professionals and organisations, and these will take time to be identified, evolve and to become embedded into routine work practices. New competencies will be required of healthcare staff, not only so they can interpret the various accounts when making decisions, but also support the validation (Combalia et al. 2022) and auditing of AI system performance and of the overall diagnostic process (Liu et al. 2022, Nix et al. 2022). It should be noted that some of the resource savings promised by the adoption of AI might be offset by the effort demanded for AI system validation and auditing.

Inherent in our approach is that issues are only understood progressively as solutions are developed and stakeholders make different judgements about the nature of the problem and the value of the solution. We aim to explore the use of co-production methodologies (Hartswood et al. 2002b, Muller et al. 2019, Voss et al. 2009) in the design and development of accounts to ensure that organisational learning is effectively supported over time as socio-material practices for accountability adapt to fit new technological affordances (Williams et al. 2005).

Finally, so far, we have only considered the delivery of trustworthy AI from the perspective of accountability *within* organisations. However, there are also inter-organisational factors that will need to be addressed. There is increasing acknowledgement of the need for standards and regulation to provide professionals and the public alike with (re)assurance of their trustworthiness (Lötsch et al. 2021). Hence, trust architectures for AI will also be shaped by industry, professional and regulatory standards and for accuracy, safety, (absence of) bias, risk, auditing, etc. (Health 2022) and frameworks for auditing compliance (Raji et al. 2020) and we aim to investigate how these are defined and are then embedded into organisational practice.

**ACKNOWLEDGEMENTS**

We would like to acknowledge the support of the UK Alan Turing Institute for Data Science and AI (Grant number EP/N510129/1) and the PathLake project, funded by Innovate UK.

**REFERENCES**

Abdul, A., Vermeulen, J., Wang, D., Lim, B. Y., & Kankanhalli, M. (2018, April). Trends and trajectories for explainable, accountable and intelligible systems: An HCI research agenda. In Proceedings of the 2018 CHI conference on human factors in computing systems (pp. 1-18).

Ackerman, S., Farchi, E., Raz, O., Zalmanovici, M., & Dube, P. (2020). Detection of data drift and outliers affecting machine learning model performance over time. arXiv preprint arXiv:2012.09258.

Adadi, A., & Berrada, M. (2018). Peeking inside the black-box: a survey on explainable artificial intelligence (XAI). IEEE access, 6, 52138-52160.

Albayram, Y., Jensen, T., Khan, M. M. H., Buck, R., & Coman, E. (2019). Investigating the Effect of System Reliability, Risk, and Role on U''rs' Emotions and Attitudes toward a Safety-Critical Drone System. International Journal of Human–Computer Interaction, 35(9), 761-772.

Alberdi, E., Povyakalo, A. A., Strigini, L., Ayton, P., Hartswood, M., Procter, R., & Slack, R. (2005). Use of computer-aided detection (CAD) tools in screening mammography: a multidisciplinary investigation. The British journal of radiology, 78(suppl_1), S31-S40.

ALI (Ada Lovelace Institute), ANI (AI Now Institute), & OGP (Open Government Partnership. (2021). Algorithmic Accountability for Public Sector: Learning from the First Wave of Policy Implementation.

Anderson, R. (1994). Representations and requirements: the value of ethnography in system design. Human-computer interaction, 9(2), 151-182.

Anderson, S., Hartswood, M., Procter, R., Rouncefield, M., Slack, R., Soutter, J., & Voss, A. (2003, September). Making autonomic computing systems accountable: the problem of human computer intera[ct]ion. In *14th International Workshop on Database and Expert Systems Applications, 2003. Proceedings.* (pp. 718-724). IEEE.

Anderson, T. (2017). *How North Sea Oil & Gas Workers Make Sense of Safety*, PhD Thesis, Lancaster University Management School.

Antoniadi, A. M., Du, Y., Guendouz, Y., Wei, L., Mazo, C., Becker, B. A., & Mooney, C. (2021). Current challenges and future opportunities for XAI in machine learning-based clinical decision support systems: a systematic review. Applied Sciences, 11(11), 5088.

Arora, A. (2020). Conceptualising artificial intelligence as a digital healthcare innovation: an introductory review. Medical Devices (Auckland, NZ), 13, 223.

Aversa, P., Cabantous, L., & Haefliger, S. (2018). When decision support systems fail: Insights for strategic information systems from Formula 1. The Journal of Strategic Information Systems, 27(3), 221-236.

Beede, E., Baylor, E., Hersch, F., Iurchenko, A., Wilcox, L., Ruamviboonsuk, P., & Vardoulakis, L. M. (2020, April). A human-centered evaluation of a deep learning system deployed in clinics for the detection of diabetic retinopathy. In *Proceedings of the 2020 CHI conference on human factors in computing systems* (pp. 1-12).

Bellotti, V., & Edwards, K. (2001). Intelligibility and accountability: human considerations in context-aware systems. *Human–Computer Interaction*, *16*(2-4), 193-212.

Bittner, E. (1965). The concept of organization. Social research. Oct 1:239-55.

Black, A. D., Car, J., Pagliari, C., Anandan, C., Cresswell, K., Bokun, T., … & Sheikh, A. (2011). The impact of eHealth on the quality and safety of health care: a systematic overview. PLoS medicine, 8(1), e1000387.

Button, G., & Harper, R. (1995). The Relevance of 'work-practice' for design. Computer Supported Cooperative Work (CSCW), 4(4), 263-280.

Button, G. & Dourish, P. (1996). Technomethodology: paradoxes and possibilities. In *CHI* (Vol. 96, pp. 19-26).

Button, G. & Sharrock, W. (1997). The production of order and the order of production: possibilities for distributed organisations, work and technology in the print industry. In Proceedings of the Fifth European Conference on Computer Supported Cooperative Work (pp. 1-16). Springer, Dordrecht.

Button, G., Crabtree, A., Rouncefield, M., & Tolmie, P. (2015). Deconstructing ethnography. Towards a social methodology for ubiquitous computing and interactive systems design. Dordrecht: Springer.

Cai, C.J., Winter, S., Steiner, D., Wilcox, L., & Terry, M. (2019a). "Hello AI": uncovering the onboarding needs of medical practitioners for human-AI collaborative decision-making. *Proceedings of the ACM on Human-computer Interaction*, *3*(CSCW), pp.1-24.

Cai, C. J., Reif, E., Hegde, N., Hipp, J., Kim, B., Smilkov, D., ... & Terry, M. (2019b). Human-centered tools for coping with imperfect algorithms during medical decision-making. In *Proceedings of the 2019 chi conference on human factors in computing systems* (pp. 1-14).

Carvalho, D. V., Pereira, E. M., & Cardoso, J. S. (2019). Machine learning interpretability: A survey on methods and metrics. Electronics, 8(8), 832.

Chatila, R., Dignum, V., Fisher, M., Giannotti, F., Morik, K., Russell, S., & Yeung, K. (2021). Trustworthy ai. In Reflections on Artificial Intelligence for Humanity (pp. 13-39). Springer, Cham.

Clarke, K., Hardstone, G., Rouncefield, M., & Sommerville, I. (Eds.). (2006). Trust in technology: A socio-technical perspective (Vol. 36). Springer Science & Business Media.

Cohen, S.L., Blanks, R.G., Jenkins, J., & Kearins, O. (2018). Role of performance metrics in breast screening imaging–where are we and where should we be? Clinical radiology, 73(4), pp.381-388.

Combalia, M., Codella, N., Rotemberg, V., Carrera, C., Dusza, S., Gutman, D., … & Malvehy, J. (2022). Validation of artificial intelligence prediction models for skin cancer diagnosis using dermoscopy images: the 2019 International Skin Imaging Collaboration Grand Challenge. The Lancet Digital Health, 4(5), e330-e339.

Coskun, E., & Grabowski, M. (2004). Impacts of User Interface Complexity on User Acceptance in Safety-Critical Systems. *AMCIS 2004 Proceedings* (2004): 432.

Coulter, J. (1983). Contingent and a priori structures in sequential analysis. *Human Studies*: 361-376.

Coulter, J. (1989). *Mind in action*. Humanities Press International, 1989.

Couture, H. D., Marron, J. S., Perou, C. M., Troester, M. A., & Niethammer, M. (2018, September). Multiple instance learning for heterogeneous images: Training a cnn for histopathology. In *International Conference on Medical Image Computing and Computer-Assisted Intervention* (pp. 254-262). Springer, Cham.

Crabtree, A., & Rodden, T. (2004). Domestic routines and design for the home. *Computer Supported Cooperative Work* 13, no. 2 (2004): 191-220.

Cummings, M. (2006). Automation and accountability in decision support system interface design.

Das, A., & Rad, P. (2020). Opportunities and challenges in explainable artificial intelligence (xai): A survey. *arXiv preprint arXiv:2006.11371*.

Davis, S.E., Lasko, T.A., Chen, G., Siew, E.D., & Matheny, M.E. (2017a). Calibration drift in regression and machine learning models for acute kidney injury. Journal of the American Medical Informatics Association, 24(6), pp.1052-1061.

Davis, S.E., Lasko, T.A., Chen, G., & Matheny, M.E. (2017b). Calibration drift among regression and machine learning models for hospital mortality. In AMIA Annual Symposium Proceedings (Vol. 2017, p. 625). American Medical Informatics Association.

Davis, S.E., Greevy Jr, R.A., Fonnesbeck, C., Lasko, T.A., Walsh, C.G., & Matheny, M.E. (2019). A nonparametric updating method to correct clinical prediction model drift. Journal of the American Medical Informatics Association, 26(12), pp.1448-1457.

Dourish, P. (1993). Culture and control in a media space. Proceedings of the European Conference on Computer-Supported Cooperative Work, ECSCW 93. Amsterdam: Kluwer.

Dourish, P. (1997). Accounting for system behaviour: Representation, reflection and Resourceful action. In M. Kyng & L. Mathiassen (Eds.), *Computers and design in con- text* (pp. 145–170). Cambridge, MA: MIT Press

Dourish, P. (2001a). Seeking a foundation for context-aware computing. *Human-Computer Interaction, 16*, 229–241.

Dourish, P. (2001b). Process descriptions as organisational accounting devices: the dual use of workflow technologies. In *Proceedings of the 2001 International ACM SIGGROUP Conference on Supporting Group Work*, pp. 52-60. 2001b.

Dourish, P. (2006). Implications for design. In Proceedings of the SIGCHI conference on Human Factors in computing systems, pp. 541-550.

Du, M., Liu, N., & Hu, X. (2019). Techniques for interpretable machine learning. Communications of the ACM, 63(1), 68-77.

Egermark, M., Blasiak, A., Remus, A., Sapanel, Y., & Ho, D. (2022). Overcoming Pilotitis in Digital Medicine at the Intersection of Data, Clinical Evidence, and Adoption. Advanced Intelligent Systems. 10.1002/aisy.202200056

Ehsan, U., Liao, Q. V., Muller, M., Riedl, M. O., & Weisz, J. D. (2021, May). Expanding explainability: towards social transparency in AI systems. In Proceedings of the 2021 CHI Conference on Human Factors in Computing Systems (pp. 1-19).

Eriksén, S. (2002). Designing for accountability. In Proceedings of the second Nordic conference on Human-computer interaction (pp. 177-186).

Fenton, N., Littlewood, B., Neil, M., Strigini, L., Sutcliffe, A., & Wright, D. (1998). Assessing dependability of safety critical systems using diverse evidence. IEE Proceedings-Software, 145(1), 35-39.

Ferri, C., Hernández-Orallo, J., & Modroiu, R. (2009). An experimental comparison of performance measures for classification. Pattern Recognition Letters, 30(1), pp.27-38.

Garfinkel, H. (1967). *Studies in Ethnomethodology*, Englewood Cliffs, Prentice-Hall.

Garfinkel, H., Lynch, M., & Livingston, E. (1981). The work of a discovering science construed with materials from the optically discovered pulsar, *Philosophy of Social Science*, 11, 131-158

Gilpin, L. H., Bau, D., Yuan, B. Z., Bajwa, A., Specter, M., & Kagal, L. (2018). Explaining Explanations: An Approach to Evaluating Interpretability of Machine Learning. *arXiv preprint arXiv:1806.00069*.

Glaser, V. L., Pollock, N., & D'Adderio, L. (2021). The biography of an algorithm: Performing algorithmic technologies in organizations. *Organization Theory*, *2*(2), 26317877211004609.

Goodwin, C. (1994). Professional vision. American Anthropologist, 96(3).

Graham, S., Minhas, F., Bilal, M., Ali, M., Tsang, Y. W., Eastwood, M., ... & Rajpoot, N. (2022). Screening of normal endoscopic large bowel biopsies with artificial intelligence: a retrospective study. *medRxiv*.

Greenhalgh, T., Wherton, J., Papoutsi, C., Lynch, J., Hughes, G., Hinder, S., … & Shaw, S. (2017). Beyond adoption: a new framework for theorizing and evaluating nonadoption, abandonment, and challenges to the scale-up, spread, and sustainability of health and care technologies. Journal of medical Internet research, 19(11), e8775.

Guan, J. (2019). Artificial intelligence in healthcare and medicine: promises, ethical challenges and governance. Chinese Medical Sciences Journal, 34(2), 76-83.

Guidotti, R., Monreale, A., Ruggieri, S., Turini, F., Giannotti, F., & Pedreschi, D. (2018). A survey of methods for explaining black box models. ACM computing surveys (CSUR), 51(5), 1-42.

Hartswood, M., Procter, R., Rouncefield, M., & Slack, R. (2002a). Performance management in breast screening: A case study of professional vision. Cognition, Technology & Work, 4(2), 91-100.

Hartswood, M., Procter, R., Slack, R., Vob, A., Buscher, M., Rouncefield, M., & Rouchy, P. (2002b). Co-realisation: Towards a principled synthesis of ethnomethodology and participatory design. Scandinavian Journal of Information Systems, 14(2), 2.

Hartswood, M., Procter, R., Rouncefield, M., Slack, R., Soutter, J., & Voss, A. (2003). 'Repairing' the Machine: A Case Study of the Evaluation of Computer-Aided Detection Tools in Breast Screening. In ECSCW 2003 (pp. 375-394). Springer, Dordrecht.

Hartswood, M., Procter, R., Rouncefield, M., & Slack, R. (2007). Cultures of Reading in Mammography. In Francis, D., & Hester, S. (Eds.), Orders of Ordinary Action: Respecifying Sociological Knowledge. Ashgate Publishing.

Heath, C., & Luff, P. (1991). Collaborative activity and technological design: Task coordination in London Underground control rooms. In Proceedings of the Second European Conference on Computer-Supported Cooperative Work ECSCW'91 (pp. 65-80). Springer, Dordrecht.

Heathfield, H. A., & Wyatt, J. (1993). Philosophies for the design and development of clinical decision-support systems. Methods of information in medicine, 32(01), 01-08.

Health, T. L. D. (2022). Holding artificial intelligence to account. The Lancet. Digital health, S2589-7500.

Henne, M., Schwaiger, A., Roscher, K., & Weiss, G. (2020, February). Benchmarking Uncertainty Estimation Methods for Deep Learning With Safety-Related Metrics. In SafeAI@ AAAI (pp. 83-90).

Henriksen, A., Enni, S., & Bechmann, A. (2021). Situated accountability: Ethical principles, certification standards, and explanation methods in applied AI. In *Proceedings of the 2021 AAAI/ACM Conference on AI, Ethics, and Society* (pp. 574-585).

Hughes, J., King, V., Rodden, T., & Andersen, H. (1994, October). Moving out from the control room: ethnography in system design. In Proceedings of the 1994 ACM conference on Computer supported cooperative work (pp. 429-439).

Jirotka, M., Procter, R., Hartswood, M., Slack, R., Simpson, A., Coopmans, C., Hinds, C., & Voss, A. (2005). Collaboration and trust in healthcare innovation: The eDiaMoND case study. Computer Supported Cooperative Work (CSCW), 14(4), pp.369-398.

Johnson, C. (2002). Software tools to support incident reporting in safety-critical systems. *Safety Science* 40, no. 9 (2002): 765-780.

Kaur, D., Uslu, S., Rittichier, K.J., & Durresi, A. (2022). Trustworthy artificial intelligence: a review. *ACM Computing Surveys (CSUR)*, *55*(2), 1-38.

Keane, P. A. & Topol, E. J. (2018). With an eye to AI and autonomous diagnosis. NPJ Digit Med 1, 40.

Knowles, B., Harding, M., Blair, L., Davies, N., Hannon, J., Rouncefield, M., & Walden, J. (2014, February). Trustworthy by design. In Proceedings of the 17[th] ACM conference on Computer supported cooperative work & social computing (pp. 1060-1071).

Knowles, B., Rouncefield, M., Harding, M., Davies, N., Blair, L., Hannon, J., Walden, J., & Wang, D. (2015, February). Models and patterns of trust. In Proceedings of the 18[th] ACM Conference on Computer Supported Cooperative Work & Social Computing (pp. 328-338).

Leslie, D. (2019). Understanding artificial intelligence ethics and safety. arXiv preprint arXiv:1906.05684. https://www.turing.ac.uk/sites/default/files/2019-06/understanding_artificial_intelligence_ethics_and_safety.pdf

Liao, Q. V., Gruen, D., & Miller, S. (2020, April). Questioning the AI: informing design practices for explainable AI user experiences. In Proceedings of the 2020 CHI Conference on Human Factors in Computing Systems (pp. 1-15).

Liu, H., Estiri, H., Wiens, J., Goldenberg, A., Saria, S., & Shah, N. (2019). AI model development and validation. Artificial Intelligence in Healthcare: The Hope, the Hype, the Promise, the Peril, 119.

Liu, X., Glocker, B., McCradden, M. M., Ghassemi, M., Denniston, A. K., & Oakden-Rayner, L. (2022). The medical algorithmic audit. The Lancet Digital Health.

Lötsch, J., Kringel, D., & Ultsch, A. (2021). Explainable artificial intelligence (XAI) in biomedicine: Making AI decisions trustworthy for physicians and patients. BioMedInformatics, 2(1), 1-17.

Luff, P., Hindmarsh, J., & Heath, C. (eds.) (2000). Workplace studies: Recovering work practice and informing system design. Cambridge University Press.

Luhmann, N. (2018). Trust and power. John Wiley & Sons.

Machlev, R., Heistrene, L., Perl, M., Levy, K. Y., Belikov, J., Mannor, S., & Levron, Y. (2022). Explainable Artificial Intelligence (XAI) techniques for energy and power systems: Review, challenges and opportunities. *Energy and AI*, 100169.

Markus, A.F., Kors, J.A., & Rijnbeek, P.R. (2021). The role of explainability in creating trustworthy artificial intelligence for health care: a comprehensive survey of the terminology, design choices, and evaluation strategies. Journal of Biomedical Informatics, 113, 103655.

Martin, D., Mariani, J., & Rouncefield, M. (2009). Practicalities of participation: Stakeholder involvement in an electronic patient records project. In Voss et al., Configuring User-Designer Relations (pp. 133-155). Springer, London.

Martin, D., & Sommerville, I. (2004). Patterns of cooperative interaction: Linking ethnomethodology and design. *ACM Transactions on Computer-Human Interaction (TOCHI)*, *11*(1), 59-89.

Martin, D., Rouncefield, M., & Sommerville, I. (2006). Patterns for dependable design. In Trust in Technology: A Socio-technical Perspective (pp. 147-168). Springer, Dordrecht.

McKinney, S.M., Sieniek, M., Godbole, V., Godwin, J., Antropova, N., Ashrafian, H., Back, T., Chesus, M., Corrado, G.S., Darzi, A., & Etemadi, M. (2020). International evaluation of an AI system for breast cancer screening. Nature, 577(7788), pp.89-94.

Mentler, T., Reuter, C., & Geisler, S. (2016). Introduction to this Special Issue on "Human-Machine Interaction and Cooperation in Safety-Critical Systems". I-com, 15(3), 219-226.

Meteier, Q., Capallera, M., Angelini, L., Mugellini, E., Khaled, O.A., Carrino, S., De Salis, E., Galland, S., & Boll, S. (2019, September). Workshop on explainable AI in automated driving: a user-centered interaction approach. In Proceedings of the 11[th] International Conference on Automotive User Interfaces and Interactive Vehicular Applications: Adjunct Proceedings (pp. 32-37).

Mittelstadt, B., Russell, C., & Wachter, S. (2019, January). Explaining explanations in AI. In Proceedings of the conference on fairness, accountability, and transparency (pp. 279-288).

Muller, M., Feinberg, M., George, T., Jackson, S. J., John, B. E., Kery, M. B., & Passi, S. (2019, May). Human-centered study of data science work practices. In Extended Abstracts of the 2019 CHI Conference on Human Factors in Computing Systems (pp. 1-8).

Musen, M. A., Middleton, B., & Greenes, R. A. (2021). Clinical decision-support systems. In Biomedical informatics (pp. 795-840). Springer, Cham.

Nix, M., Onisiforou, G., & Painter, S. (2022). Understanding healthcare workers confidence in AI. NHS AI Lab & Health Education England. Available at: https://digital-transformation.hee.nhs.uk/binaries/content/assets/digital-transformation/dart-ed/understandingconfidenceinai-may22.pdf

Oakden-Rayner, L., Gale, W., Bonham, T. A., Lungren, M. P., Carneiro, G., Bradley, A. P., & Palmer, L. J. (2022). Validation and algorithmic audit of a deep learning system for the detection of proximal femoral fractures in patients in the emergency department: a diagnostic accuracy study. *The Lancet Digital Health*, *4*(5), e351-e358.

Pedreschi, D., Giannotti, F., Guidotti, R., Monreale, A., Pappalardo, L., Ruggieri, S., & Turini, F. (2018). Open the black box data-driven explanation of black box decision systems. arXiv preprint arXiv:1806.09936.

Pocevičiūtė, M., Eilertsen, G., & Lundström, C. (2020). Survey of XAI in digital pathology. In *Artificial intelligence and machine learning for digital pathology* (pp. 56-88). Springer, Cham.

Procter, R., Rouncefield, M., Balka, E., & Berg, M. (2006). CSCW and dependable healthcare systems. Computer Supported Cooperative Work (CSCW), 15(5-6), 413-418.

Procter, R., Tolmie, P., & Rouncefield, M. (2022). Trust, Professional Vision and Diagnostic Work. In Ontika, N. N., Sasmannshausen, S. M., Syed, H. A., & de Carvalho, A. F. P. (Eds.) Exploring Human-Centered AI in Healthcare: A Workshop Report. International Reports on Socio-Informatics, vol. 19, no. 2.

Raji, I. D., Smart, A., White, R. N., Mitchell, M., Gebru, T., Hutchinson, B., … & Barnes, P. (2020, January). Closing the AI accountability gap: Defining an end-to-end framework for internal algorithmic auditing. In Proceedings of the 2020 conference on fairness, accountability, and transparency (pp. 33-44).

Rakha, E. A., Soria, D., Green, A. R., Lemetre, C., Powe, D. G., Nolan, C. C., ... & Ellis, I. O. (2014). Nottingham Prognostic Index Plus (NPI+): a modern clinical decision making tool in breast cancer. British journal of cancer, 110(7), 1688-1697.

Randall, D., Harper, R., Rouncefield, M. (2005). Fieldwork and Ethnography: A perspective from CSCW. In EPIC 2005: Ethnographic Praxis in Industry Conference Proceedings, Redmond, USA. Vol. 2005, No. 1. Oxford, UK: Blackwell Publishing Ltd. Pp. 81–99.

Randall, D., Harper, R., & Rouncefield, M. (2007). Fieldwork for design: theory and practice. Springer Science & Business Media.

Randall, D. (2018). Investigation and design. Socio-Informatics: a practice-based perspective on the design and use of IT artifacts (1st ed). Oxford University Press, Oxford, UK, 221-241.

Richardson, J.P., Smith, C., Curtis, S., Watson, S., Zhu, X., Barry, B., & Sharp, R. (2021). Patient apprehensions about the use of artificial intelligence in healthcare. *NPJ digital medicine*, *4*(1), 1-6.

Riegelsberger, J., Sasse, M. A., & McCarthy, J. D. (2005). The mechanics of trust: A framework for research and design. International Journal of Human-Computer Studies, 62(3), 381-422.

Sacks, H., Schegloff, E., & Jefferson, G. (1978). A simplest systematics for the organization of turn taking for conversation. Studies in the organization of conversational interaction. Academic Press. 7-55.

Sacks, H. (1992). *Lectures on Conversation, Volumes I & II*, (edited by G. Jefferson), Malden, MA: Blackwell.

Sanneman, L., & Shah, J. A. (2022). The Situation Awareness Framework for Explainable AI (SAFE-AI) and Human Factors Considerations for XAI Systems. *International Journal of Human–Computer Interaction*, 1-17.

Saarela, M., & Geogieva, L. (2022). Robustness, Stability, and Fidelity of Explanations for a Deep Skin Cancer Classification Model. *Applied Sciences*, *12*(19), 9545.

Sellen, A., Harper, R. (2003). *The myth of the paperless office*. MIT press.

Shneiderman, B. (2020). Bridging the gap between ethics and practice: guidelines for reliable, safe, and trustworthy human-centered AI systems. ACM Transactions on Interactive Intelligent Systems (TiiS), 10(4), pp.1-31.

Singh, A., Sengupta, S., & Lakshminarayanan, V. (2020). Explainable deep learning models in medical image analysis. Journal of Imaging, 6(6), 52.

Slack, R. S., Procter, R., Hartswood, M., Voss, A., & Rouncefield, M. (2010). Suspicious minds?. In Ethnographies of Diagnostic Work. In Buscher, M., Goodwin, D., & Mesman, J. (Eds.) Ethnographies of Diagnostic Work. Palgrave Press.

Smith, H. (2021). Clinical AI: opacity, accountability, responsibility and liability. *AI & SOCIETY*, *36*(2), 535-545.

Soria, D., Garibaldi, J. M., Ambrogi, F., Green, A. R., Powe, D., Rakha, E., ... & Ellis, I. O. (2010). A methodology to identify consensus classes from clustering algorithms applied to immunohistochemical data from breast cancer patients. Computers in biology and medicine, 40(3), 318-330.

Suchman, Lucy A. (1987). *Plans and situated actions: The problem of human-machine communication*. Cambridge University Press.

Suchman, L., Trigg, R., & Blomberg, J. (2002). Working artefacts: ethnomethods of the prototype. The British journal of sociology, 53(2), 163-179.

Tolmie, P., Pycock, J., Diggins, T., MacLean, A., & Karsenty, A. (2002, April). Unremarkable computing. In Proceedings of the SIGCHI conference on Human factors in computing systems (pp. 399-406).

Tolmie, P., & Rouncefield, M. (2016). Organizational acumen. In *Ethnomethodology at Work*, pp. 63-82. Routledge.

Victorelli, E. Z., Dos Reis, J. C., Hornung, H., & Prado, A. B. (2020). Understanding human-data interaction: Literature review and recommendations for design. International Journal of Human-Computer Studies, 134, 13-32.

Voss, A., Procter, R., Slack, R., Hartswood, M., & Rouncefield, M. (2009). Design as and for collaboration: Making sense of and supporting practical action. In Configuring user-designer relations (pp. 31-58). Springer, London.

Wang, D., Yang, Q., Abdul, A., & Lim, B. Y. (2019, May). Designing theory-driven user-centric explainable AI. In Proceedings of the 2019 CHI conference on human factors in computing systems (pp. 1-15).

Watson, R. (2009). Constitutive practices and Garfinkel's notion of trust: Revisited. *Journal of Classical Sociology*, *9*(4), 475-499.

Williams, R., Stewart, J., & Slack, R. (2005). *Social learning in technological innovation: Experimenting with information and communication technologies*. Edward Elgar Publishing.

Wu, H., Chen, W., Xu, S., & Xu, B. (2021). Counterfactual supporting facts extraction for explainable medical record based diagnosis with graph network. In *Proceedings of the 2021 Conference of the North American Chapter of the Association for Computational Linguistics: Human Language Technologies* (pp. 1942-1955).

Yang, Q., Steinfeld, A., & Zimmerman, J. (2019, May). Unremarkable ai: Fitting intelligent decision support into critical, clinical decision-making processes. In Proceedings of the 2019 CHI Conference on Human Factors in Computing Systems (pp. 1-11).

Yang, G., Ye, Q., & Xia, J. (2022). Unbox the black-box for the medical explainable ai via multi-modal and multi-centre data fusion: A mini-review, two showcases and beyond. Information Fusion, 77, 29-52.

Zhang, Y., Weng, Y., & Lund, J. (2022). Applications of Explainable Artificial Intelligence in Diagnosis and Surgery. *Diagnostics*, *12*(2), 237.

Zimmerman, D. (1971). The Practicalities of Rule Use. In J.D. Douglas (ed.), *Understanding Everyday Life: Toward the Reconstruction of Sociological Knowledge*, London: Routledge and Kegan Paul.